%% file: main.tex
\documentclass[letterpaper,10pt,journal]{IEEEtran}

\newif\ifanonymous
\anonymousfalse

\usepackage[linesnumbered,ruled,vlined]{algorithm2e}

\usepackage{xcolor}
\definecolor{darkred}{RGB}{205,38,38}
\definecolor{darkblue}{RGB}{0,0,128}

\usepackage[caption=false,font=footnotesize]{subfig}
\usepackage{capt-of}
\usepackage{float} 
 
\usepackage{enumitem}
\usepackage{booktabs} 
\usepackage{multirow}
\usepackage{cite}
\usepackage{xspace}
\usepackage{array}
\newcolumntype{L}[1]{>{\raggedright\let\newline\\\arraybackslash\hspace{0pt}}m{#1}}
\newcolumntype{C}[1]{>{\centering\let\newline\\\arraybackslash\hspace{0pt}}m{#1}}
\newcolumntype{R}[1]{>{\raggedleft\let\newline\\\arraybackslash\hspace{0pt}}m{#1}}
\usepackage{threeparttable}
\usepackage{makecell}
\usepackage{balance}
\usepackage{amsmath}
\usepackage{amssymb,bm}
\usepackage{amsfonts}
\usepackage{graphicx}
\usepackage{pifont}

\usepackage[colorlinks=true,
linkcolor=darkred,       
anchorcolor=blue, 
citecolor=darkblue,        
filecolor=darkred, 
urlcolor=blue,
]{hyperref}

\newcommand{\methodname}{RAPID\xspace}
\newtheorem{theorem}{Theorem}

\title{\methodname: Architecture-Agnostic Post-Training for \\Real-World Humanoid Manipulation}

\ifanonymous
\author{Anonymous Submission}
\else
\author{
Kun~Song$^{1,2}$, Luyao~Zhang$^{2,3}$, Wenqi~Ge$^{1,2}$, Yin~Zhang$^{1,2}$, Zejia~Liu$^{1}$,\\
Tengye~Xu$^{1,2}$, Dongjie~Yu$^{1}$, Zhen~Fu$^{2}$, Hua~Chen$^{2}$, and Jia~Pan$^{1}$
\thanks{$^{1}$The University of Hong Kong. $^{2}$LimX Dynamics Technology Co., Ltd. $^{3}$Delft University of Technology.}
}
\fi

\begin{document}

\IEEEaftertitletext{%
    \begin{minipage}{\textwidth}
        \centering
        \includegraphics[width=\textwidth]{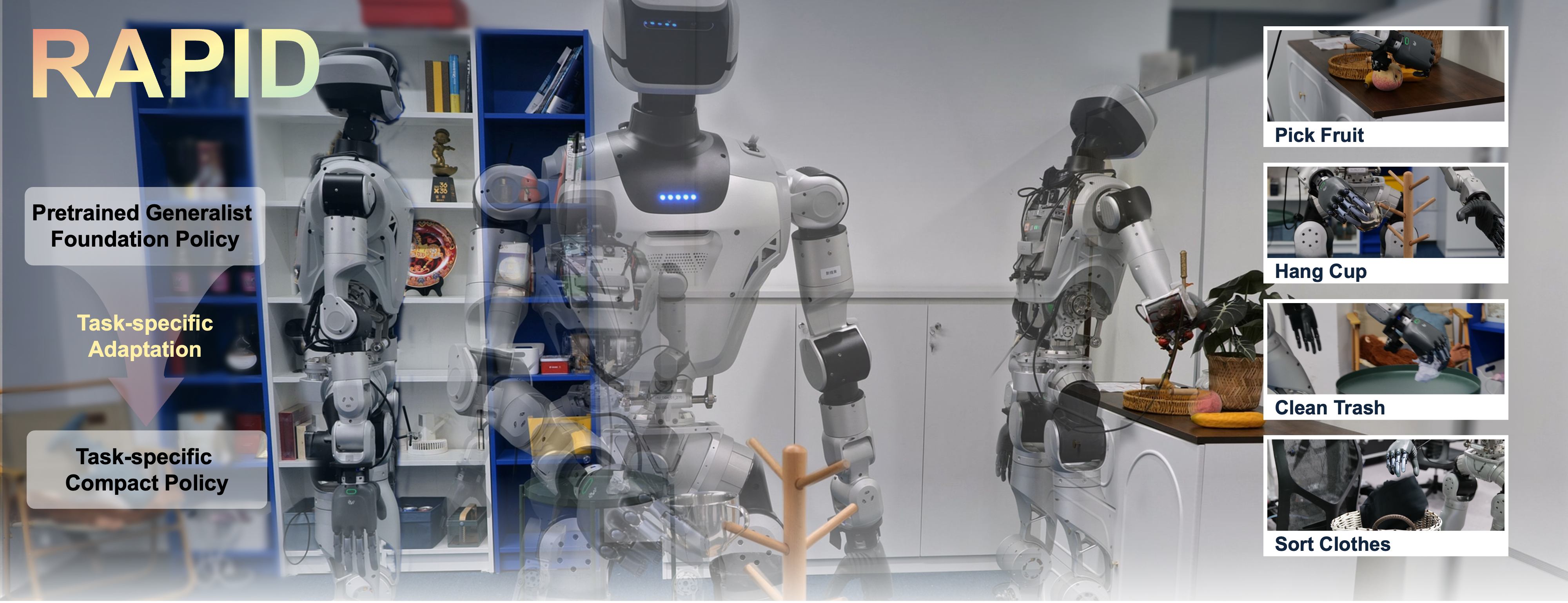}
        \captionof{figure}{Real-world humanoid policy adaptation with \methodname on \textit{Pick Fruit}, \textit{Hang Cup}, \textit{Clean Trash}, and \textit{Sort Clothes}. \methodname focuses on transforming a teacher policy, regardless of its architecture, into a task-specific student policy with a high success rate through distillation and online reinforcement learning.}
        \label{fig:overview}
    \end{minipage}
    \par\vspace{1em}
}

\maketitle

\begin{abstract}
Humanoid robots promise versatile deployment across domestic and industrial settings by unifying legged mobility with bimanual dexterity.
While pretrained vision--language--action (VLA) models, world action models (WAMs), and agentic policies provide broad visuomotor priors, their performance degrades sharply under real-world distribution shifts.
Yet adapting foundation policies on physical hardware remains bottlenecked by prohibitive fine-tuning compute, inaccessible weights in proprietary models, and non-differentiable hybrid architectures.
We present \methodname, an architecture-agnostic post-training framework that transforms arbitrary black-box teacher policies into compact, high-performing whole-body student controllers without accessing teacher parameters or internal features.
Grounded in a theoretical adaptation bound, \methodname unifies three synergistic stages:
offline distillation establishes safe behavioral priors;
on-policy distillation (OPD) with teacher takeover and counterfactual masked-action alignment eliminates compounding state-distribution shift; and
Intervention-guided Policy Refinement (IPR) couples chunk-level implicit Q-learning (IQL) with jump-free whole-body human corrections to surpass the teacher ceiling.
With approximately one hour of post-training per task on a single consumer GPU, \methodname elevates the mean success rate across four real-world humanoid tasks from 20.51\% to 89.72\%, reduces deployment compute by 98.68\%, and generalizes seamlessly across VLA, WAM, and agentic teachers.
Project website: \url{https://song-kun.github.io/RAPID/}.
\end{abstract}


\input{sec/introduction}
\input{sec/related_works}
\input{sec/method}
\input{sec/experiments}
\input{sec/conclusion}

\bibliographystyle{IEEEtran-first-et-al-no-url}
\bibliography{ref}

\raggedbottom
\appendix
\input{sec/theory_appendix}
\subsection{Training Hyperparameters}

\begin{table}[!t]
    \caption{Optimization hyperparameters for the base policies, baselines, and \methodname training stages.}
    \label{tab:training_hyperparameters}
    \centering
    \footnotesize
    \renewcommand{\arraystretch}{1.05}
    \setlength{\tabcolsep}{2pt}
    \begin{tabular}{@{}L{\dimexpr0.47\columnwidth-2\tabcolsep\relax}
        C{\dimexpr0.265\columnwidth-\tabcolsep\relax}
        C{\dimexpr0.265\columnwidth-\tabcolsep\relax}@{}}
        \toprule
        \textbf{Hyperparameter}
        & \textbf{\textit{Pick Fruit}} & \textbf{\textit{Hang Cup}} \\
        \midrule
        \multicolumn{3}{@{}l}{\textbf{GR00T base policy}} \\
        Batch size & 64 & 64 \\
        Learning rate / weight decay
        & \multicolumn{2}{c}{$1\times10^{-4}$ / $1\times10^{-5}$} \\
        Optimization steps & 30,000 & 20,000 \\
        \midrule
        \multicolumn{3}{@{}l}{\textbf{HG-DAgger}~\cite{kelly2019hg}} \\
        Batch size / learning rate
        & \multicolumn{2}{c}{64 / $1\times10^{-5}$} \\
        Optimization steps & \multicolumn{2}{c}{10,000}\\
        Corrective frames & 10,761 & 12,526 \\
        Learning-rate schedule & \multicolumn{2}{c}{Cosine, 5\% warm-up}  \\
        \midrule
        \multicolumn{3}{@{}l}{\textbf{DSRL}~\cite{wagenmaker2025steering}} \\
        Batch size & \multicolumn{2}{c}{128} \\
        Actor / critic / temperature learning rate
        & \multicolumn{2}{c}{$3\times10^{-5}$ / $3\times10^{-5}$ / $3\times10^{-5}$} \\
        $\gamma$ / UTD ratio / number of critics
        & \multicolumn{2}{c}{0.99 / 1 / 2} \\
        \midrule
        \multicolumn{3}{@{}l}{\textbf{RL Token}~\cite{xu2026rl}} \\
        Batch size & \multicolumn{2}{c}{128} \\
        Actor / critic learning rate
        & \multicolumn{2}{c}{$5\times10^{-5}$ / $3\times10^{-5}$} \\
        $\beta$ / actor warm-up & \multicolumn{2}{c}{5 / 30 episodes} \\
        $\gamma$ / UTD ratio / number of critics
        & \multicolumn{2}{c}{0.99 / 1 / 2} \\
        \midrule
        \multicolumn{3}{@{}l}{\textbf{\methodname Warm up}} \\
        Batch size / learning rate / weight decay
        & \multicolumn{2}{c}{32 / $1\times10^{-5}$ / $1\times10^{-4}$} \\
        Training epochs & \multicolumn{2}{c}{20} \\
        \midrule
        \multicolumn{3}{@{}l}{\textbf{\methodname OPD}} \\
        Batch size / update epochs per rollout & \multicolumn{2}{c}{64 / 3} \\
        Offline ratio / gradient clipping & \multicolumn{2}{c}{0.3 / 1.0} \\
        \midrule
        \multicolumn{3}{@{}l}{\textbf{\methodname IQL}} \\
        Batch size & \multicolumn{2}{c}{128} \\
        Actor / critic / value learning rate
        & \multicolumn{2}{c}{$5\times10^{-5}$ / $3\times10^{-5}$ / $3\times10^{-5}$} \\
        $\gamma$ / UTD ratio / number of critics
        & \multicolumn{2}{c}{0.99 / 1 / 2} \\
        IQL expectile $\tau$ & \multicolumn{2}{c}{0.7} \\
        AWR $\beta$ / weight clip & \multicolumn{2}{c}{10 / 3} \\
        \bottomrule
    \end{tabular}
\end{table}

All four tasks use $224\times224$-pixel RGB inputs, a 45-dimensional state, and a one-step observation history. The compact policy uses 16 visual tokens per camera and one state token, with proprioceptive dropout set to 0.8. It predicts 16-step chunks of 49-dimensional actions and executes 10 steps per chunk at 30~Hz.

The warm-up rollouts below are collected by the frozen teacher for offline student distillation and count toward the post-training interaction budget.

For \textit{Pick Fruit}, the policy uses a single head RGB camera, and episodes are limited to 600 steps. Post-training comprises 20 warm-up rollouts, 36 OPD rollouts, and 64 IQL rollouts.
For \textit{Hang Cup}, the policy uses a single head RGB camera, and episodes are limited to 1,000 steps. Post-training comprises 20 warm-up rollouts, 20 OPD rollouts, and 60 IQL rollouts.
For \textit{Clean Trash}, the policy uses one head RGB camera and one right wrist RGB camera. Episodes are limited to 1,000 steps, with 20 warm-up rollouts, 20 OPD rollouts, and 60 IQL rollouts for post-training.

For \textit{Sort Clothes}, the policy uses one head RGB camera and one wrist RGB camera. Post-training comprises 20 warm-up rollouts, 20 OPD rollouts, and 41 IQL rollouts.

Table~\ref{tab:training_hyperparameters} summarizes the optimization hyperparameters for the base policies, baselines, and training stages.
For DSRL, RL Token, and \methodname IQL, the target-network soft-update coefficient is 0.005.

\end{document}

%% file: sec/introduction.tex
\section{Introduction}

Humanoid robots hold broad potential across industrial manufacturing and domestic service by unifying legged mobility with bimanual dexterity in a human-compatible morphology~\cite{liang2026vivo,jiang2026wholebodyvla,cheng2024expressive,barreiros2025wholebody,kaelbling2020foundation}.
Recent progress in pretrained foundation models, like VLA models \cite{jiang2026wholebodyvla,wei2026psi0,bjorck2025gr00t}, WAMs \cite{agarwal2026cosmos}, and agentic policies \cite{jia2026agent}, provides broad semantic and visuomotor priors that map multimodal observations and language instructions into robot commands.
Nevertheless, even after internet-level and fleet-scale pretraining, generalist humanoid policies suffer sharp performance degradation when deployed in real-world target environments~\cite{huang2025improving}.
Out-of-distribution (OOD) spatial layouts, novel object geometries, egocentric viewpoint shifts, and contact-rich interaction dynamics break zero-shot execution, making rapid post-training essential for reliable physical deployment~\cite{intelligence2025pi,luo2025precise,xiao2026rove}.

However, post-training can be challenging or even infeasible during real-world deployment.
First, end-to-end fine-tuning of billion-parameter VLAs and video-diffusion WAMs demands datacenter-grade GPUs and hours of offline optimization, precluding rapid in-situ adaptation.
Second, proprietary foundation policies restrict access to model weights, intermediate activations, and internal noise schedules, ruling out both gradient-based fine-tuning and white-box latent steering~\cite{dodeja2026life}.
Third, emerging hybrid architectures, such as tool-augmented agentic systems, modular learning and control stacks, embed discrete symbolic reasoning and non-differentiable solvers that break end-to-end backpropagation and force bespoke, model-specific engineering~\cite{tagliabue2024efficient}.

To overcome these barriers simultaneously, we introduce \textbf{\methodname}, an \underline{a}rchitecture-agnostic \underline{p}ost-training framework for \underline{r}eal-world humano\underline{i}d loco-manipulation via black-box \underline{d}istillation and online reinforcement learning (RL) (Fig.~\ref{fig:overview}).
Operating strictly through action-output queries without accessing teacher parameters or internal features, \methodname transforms arbitrary low-performing foundation teachers into compact, high-performance student policies within roughly one hour of physical interaction.
Grounded in a theoretical decomposition of policy adaptation shortfall into action-fitting error, state-distribution mismatch, and reference suboptimality, \methodname unifies three synergistic stages.
Offline distillation first initializes the student from trajectories collected by the teacher.
Second, embodied on-policy distillation (OPD)~\cite{agarwal2024onpolicy} queries the frozen teacher along student-visited states---augmented with teacher-takeover horizon completion, and counterfactual masked-action alignment---to eliminate compounding rollout drift.
Third, Intervention-guided Policy Refinement (IPR) integrates chunk-level implicit Q-learning (IQL)~\cite{kostrikov2022iql,li2025qchunking} with jump-free whole-body human corrections, inducing an implicit value-tilted reference that drives task mastery well beyond the teacher ceiling.
Our main contributions are:

\begin{itemize}[leftmargin=*]
    \item We propose \textbf{\methodname}, an architecture-agnostic adaptation framework that requires zero access to teacher parameters or internal features and produces a lightweight student for independent deployment.
    \item We design an OPD mechanism uniting teacher takeover and counterfactual masked-action alignment to resolve compounding rollout errors, paired with a jump-free delta intervention scheme and Q-chunk IQL for whole-body credit assignment.
    \item We validate \methodname through real-world humanoid experiments and ablation studies, demonstrating its adaptation effectiveness and computational efficiency during training and deployment.
\end{itemize}

%% file: sec/related_works.tex
\section{Related Work}

\subsection{Pretrained Foundation Models for Manipulation}

Pretrained policies provide generalist manipulation capabilities in open-world environments.
The $\pi_0$ model couples a pretrained vision--language backbone with a flow-matching action expert, while $\pi_{0.5}$ uses heterogeneous robot and non-robot data to improve generalization to new environments~\cite{black2024pi0,intelligence2025pi05}.
WAMs additionally connect action generation to the prediction of future visual states.
WorldVLA jointly learns action prediction and action-conditioned image prediction in an autoregressive token model~\cite{cen2025worldvla}.
DreamZero instead uses a pretrained video-diffusion backbone to jointly generate future video and actions, demonstrating generalization to unseen tasks and environments on real robots~\cite{ye2026dreamzero}.
Agent as Policy (AGP) places planning and execution under a general-purpose agent that generates executable programs and revises robot actions using physical feedback~\cite{jia2026agent}.
Together, VLA models, WAMs, and agentic systems offer complementary ways to translate pretrained knowledge into robot behavior.

Extending foundation policies to humanoids requires learning coordinated behavior across the body, arms, and hands from data that cover diverse interactions.
GR00T N1 combines real robot demonstrations, simulated trajectories, generated videos, and human videos to learn manipulation priors that support data-efficient supervised adaptation on humanoid tasks~\cite{bjorck2025gr00t}.
For humanoid loco-manipulation, $\Psi_0$ separates vision--language pretraining on egocentric human video from flow-based action-expert training on humanoid trajectories, followed by task-specific adaptation for long-horizon real-world behavior~\cite{wei2026psi0}.
Besides, many works show that human and robot data can provide reusable priors and reduce downstream demonstration requirements \cite{xu2026towards, jiang2026wholebodyvla,chen2026unit}.
However, due to the distribution shift in real-world, the performance of a pretrained policy can be degraded during deployment.

\subsection{Post-Training of Manipulation Policies}

Post-training adapts manipulation policies to target tasks through corrective demonstrations, reward feedback, or both~\cite{yu2026learningfleetexperiencecollective,liang2026robometer,yang2026rise,chen2026clift,luo2024rlif}.
HG-DAgger aggregates human corrections at failure states\cite{kelly2019hg}.
HIL-SERL and RL-100 use RL-based method to update policies~\cite{luo2025precise,lei2026performant}.
RLDG improves generalist policies by fine-tuning them on high-quality data generated by task-specific RL policies~\cite{xu2025rldg}.
For generalist VLAs, RECAP integrates demonstrations, autonomous experience, and human interventions through advantage-conditioned policy learning, with related adaptation explored on humanoids~\cite{intelligence2025pi,siso2026closing}.
ROVE adapts humanoid VLAs from imperfect interventions by updating action and value heads while freezing the pretrained vision--language backbone~\cite{xiao2026rove}.
Sirius learns from autonomous rollouts and human interventions collected during deployment through behavioral cloning (BC) weighted by estimated human trust~\cite{liu2025sirius}.

Post-training has also been explored for WAMs and agentic policies.
$R^2$-WAM first repairs inconsistencies between input actions and predicted video.
The rejection stage uses negative fine-tuning to suppress actions predicted to make less task progress than demonstrated actions, improving success without additional environment interaction~\cite{xu2026r}.
RACaP distills reasoning-and-acting decision traces into the policy through rejection-sampled fine-tuning, retaining successful and executable teacher decisions~\cite{li2026racap}.

To reduce optimization costs, lightweight approaches keep pretrained policies frozen and learn auxiliary components.
DSRL optimizes noise-space steering, whereas RL Token trains a small actor--critic using VLA features and reference actions~\cite{wagenmaker2025steering,xu2026rl}.
HAF and HOIST explore noise-space adaptation for humanoid manipulation~\cite{gu2026haf,liu2026hoist}.
These methods retain the pretrained policy during inference, and feature- or noise-based variants require model-specific interfaces.
RPD combines PPO with action-level distillation from a VLA teacher to learn compact task-specific policies that run independently of the teacher~\cite{julg2025refined}.
Table~\ref{tab:post_training_comparison} compares representative approaches in terms of human involvement, training efficiency, deployment, and teacher access.

\begin{figure*}[t]
    \centering
    \includegraphics[width=\textwidth]{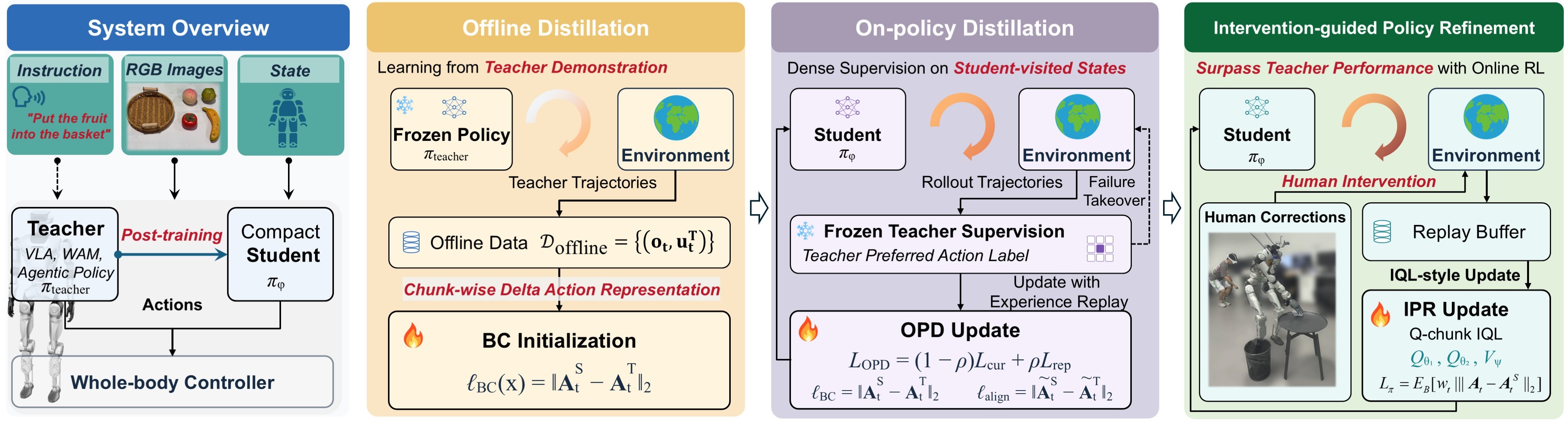}
    \caption{Overview of the \methodname framework: (1) offline distillation initializes the student from frozen teacher trajectories; (2) OPD provides teacher supervision at student-visited states; and (3) IPR improves the student using real-world rewards and human corrections.}
    \label{fig:framework}
\end{figure*}

These approaches motivate post-training that accommodates limited computation, restricted parameter access, and heterogeneous policy architectures.
\methodname follows a general architecture-agnostic formulation, learning from a frozen teacher's action outputs without accessing its parameters or modifying its architecture.
Offline distillation and OPD transfer behavior into a compact student, while IPR uses real-world interaction and human corrections for further improvement.
The resulting student runs independently of the teacher during deployment.
Fig.~\ref{fig:framework} presents the \methodname framework.

\begin{table}[!t]
    \caption{Comparison of policy post-training approaches. HIL denotes online human corrective intervention.}
    \label{tab:post_training_comparison}
    \centering
    \footnotesize
    \renewcommand{\arraystretch}{1.2}
    \setlength{\tabcolsep}{2pt}
    \begin{tabular*}{\columnwidth}{@{\extracolsep{\fill}} l c c c c c @{}}
        \toprule
        \textbf{Method}
        & \textbf{HIL}
        & \makecell{\textbf{Training}\\\textbf{Efficiency}}
        & \makecell{\textbf{Compact}\\\textbf{Deploy.}}
        & \makecell{\textbf{Base-free}\\\textbf{Inference}}
        & \makecell{\textbf{Architecture-}\\\textbf{agnostic}} \\
        \midrule
        HG-DAgger~\cite{kelly2019hg}
        & \ding{51} & \ding{55} & \ding{55} & \ding{55} & \ding{55} \\
        RPD~\cite{julg2025refined}
        & \ding{55} & \ding{55} & \ding{51} & \ding{51} & \ding{55} \\
        DSRL~\cite{wagenmaker2025steering}
        & \ding{55} & \ding{51} & \ding{55} & \ding{55} & \ding{55} \\
        RECAP~\cite{intelligence2025pi}
        & \ding{51} & \ding{55} & \ding{55} & \ding{55} & \ding{55} \\
        RL Token~\cite{xu2026rl}
        & \ding{51} & \ding{51} & \ding{55} & \ding{55} & \ding{55} \\
        \midrule
        \textbf{\methodname (Ours)}
        & \ding{51} & \ding{51} & \ding{51} & \ding{51} & \ding{51} \\
        \bottomrule
    \end{tabular*}
\end{table}

%% file: sec/method.tex
\section{Preliminaries and Problem Formulation}

\subsection{Whole-Body Controller}

We employ a humanoid whole-body controller that maps task-space commands to joint-space targets.
All anchor poses are expressed in the coordinate frame obtained by projecting the robot's base frame onto the ground plane.
At each control step, the controller receives desired anchor poses $\mathbf{q}^{i}_t\in\mathrm{SE}(3), i\in \mathcal{K}$ in this frame, forward linear velocity $v_t$, and turning angular velocity $\omega_t$.
We define the action space over the ordered anchor set $\mathcal{K}=(\mathrm{base},\mathrm{torso},\mathrm{head}, \mathrm{left\ hand},\mathrm{right\ hand})$.
The binary hand command $\mathbf{c}^{\mathrm{hand}}_t\in\{0,1\}^{2}$ specifies whether each hand is open or closed.
The whole-body controller resolves these high-level commands and outputs targets for all joints at 50~Hz.
A low-level proportional--derivative (PD) controller receives these desired joint angles and outputs joint torques at 1,000~Hz.

\subsection{Policies for Whole-Body Manipulation}

We first introduce the inputs and outputs of the teacher and student policies for humanoid whole-body manipulation.
Let $\pi_{\mathrm{teacher}}$ denote the pretrained teacher policy and $\pi_\phi$ the task-specific student policy parameterized by $\phi$.
At time step $t$, both policies receive the multimodal observation
\begin{equation}
    \mathbf{o}_t = \left(\mathbf{I}_t,\mathbf{j}^{\mathrm{body}}_{t},
    \mathbf{j}^{\mathrm{hand}}_t,\mathbf{c}^{\mathrm{hand}}_{t-1}\right),
\end{equation}
where $\mathbf{I}_t\in\mathbb{R}^{N_{\mathrm{cam}}\times3\times224\times224}$ stacks the RGB images from the $N_{\mathrm{cam}}\ge1$ cameras, with each image having resolution $224\times224$.
The remaining components are $\mathbf{j}^{\mathrm{body}}_{t} \in\mathbb{R}^{31}$, which contains the whole-body joint angles; $\mathbf{j}^{\mathrm{hand}}_t\in\mathbb{R}^{12}$, which contains the six measured joint angles of each dexterous hand; and $\mathbf{c}^{\mathrm{hand}}_{t-1}\in\{0,1\}^{2}$, the last binary open--close command.
The teacher additionally receives a task instruction $\ell$, while the student uses only $\mathbf{o}_t$ because each student is trained for a single task with a fixed instruction.

The absolute action space is defined as
\begin{equation}
    \mathbf{u}_t = \operatorname{concat}\!\left(
    \{\psi(\mathbf{q}^{i}_t)\}_{i\in\mathcal{K}},v_t,\omega_t,
    \mathbf{c}^{\mathrm{hand}}_t\right)\in\mathbb{R}^{49},
\end{equation}
where $\psi:\mathrm{SE}(3)\rightarrow\mathbb{R}^{9}$ encodes each pose $\mathbf{q}=(\mathbf{R},\mathbf{p})$ as
\begin{equation}
    \psi(\mathbf{q})
    = \left[\mathbf{p}^{\top},\mathbf{R}_{1,:},\mathbf{R}_{2,:}\right]^{\top}.
\end{equation}
Here, $\mathbf{p}\in\mathbb{R}^{3}$ is the translation vector, and $\mathbf{R}_{1,:}$ and $\mathbf{R}_{2,:}$ are the first two rows of the rotation matrix.

We denote an $H$-step absolute action chunk by
$
    \mathbf{U}_t = (\mathbf{u}_t,\ldots,\mathbf{u}_{t+H-1}).
$
We use a chunk-wise delta representation with the last issued command $\mathbf{u}_{t-1}$ as the shared reference \cite{feng2026demystifying}.
For the continuous command dimensions, each action within this chunk is encoded as
\begin{equation}
    \mathbf{a}_{t+k} = \mathbf{u}_{t+k}-\mathbf{u}_{t-1},
    \quad k=0,\ldots,H-1.
    \label{eq:delta_action}
\end{equation}
Here, $\mathbf{a}_{t+k}$ is defined within the chunk starting at $t$, whose reference remains fixed across all $H$ entries.
The two binary hand commands retain their absolute values $\mathbf{c}^{\mathrm{hand}}_{t+k}$ in $\mathbf{a}_{t+k}$ and are excluded from differencing.
These encoded actions form the delta-action chunk
$
    \mathbf{A}_t = (\mathbf{a}_t,\ldots,\mathbf{a}_{t+H-1}).
$
For training, we standardize these chunks and retain $\mathbf{A}_t$ and $\mathbf{a}_{t+k}$ for the standardized chunks and their entries.
The teacher outputs an absolute action chunk $\mathbf{U}^{T}_t\sim\pi_{\mathrm{teacher}}(\cdot\mid\mathbf{o}_t,\ell)$, which is encoded and standardized to obtain the supervision target $\mathbf{A}^{T}_t$.
The student directly predicts a standardized delta-action chunk $\mathbf{A}^{S}_t=\pi_\phi(\mathbf{o}_t)$.
At deployment, standardization is reversed, and each continuous command is reconstructed by adding its predicted delta to the same reference $\mathbf{u}_{t-1}$.
Predicted hand commands are decoded as absolute open--close commands.

\subsection{Whole-Body Teleoperation}

We collect demonstrations using a virtual reality based teleoperation system by PICO 4 \cite{darvish2023teleoperation}.
The system tracks poses of the operator's base, torso, left hand, right hand, and head to the corresponding robot body frames.
Then, a retarget system is used to transform these poses for the humanoid robots.
Joystick deflection signals from the PICO~4 handheld controllers are mapped to the robot's forward linear velocity $v_t$ and turning angular velocity $\omega_t$.
The same controllers provide switch inputs for controlling the dexterous hands: each hand command is represented as a binary open or closed state, whose corresponding hand poses are predefined.
The resulting teleoperated trajectories are recorded at 30~Hz.

\section{The \methodname System}
\label{sec:rapid_system}

As shown in Fig.~\ref{fig:framework}, \methodname converts a pretrained teacher policy into a task-adapted student through three successive stages.
In the first stage, we deploy $\pi_{\mathrm{teacher}}$ in the target environment and collect a teacher rollout dataset $\mathcal{D}_{\mathrm{offline}}$ for student distillation.
We use these trajectories to initialize the student through BC before online learning begins.

In the second stage, the student collects a trajectory with teacher takeover when a human operator signals failure.
After collection, the frozen teacher supplies action targets for recorded student states, and the student is updated using the completed rollout and replay of the initial offline teacher data.
To extract more supervision from each interaction, we also match teacher and student action predictions under the same masked observation.
This provides additional teacher targets without requiring extra environment transitions.

The distilled policy can nevertheless retain task-level failure modes that are not resolved by action imitation alone.
In the final IPR stage, we combine whole-body intervention and IQL-style \cite{kostrikov2022iql} update to further improve the success rate.

\subsection{Policy Adaptation Performance Bound Analysis}
Before the detailed implementation of \methodname, we analyze performance bound for post-training from a frozen teacher.
We define the student as $\pi_\phi$, the frozen teacher as $\pi_{\mathrm{teacher}}$.
$q$ specifies where supervision is available for offline distillation.
Following interactive imitation learning~\cite{ross2011reduction} and its continuous-action performance analysis~\cite{pan2018agile}, we model post-training as a finite-horizon decision process with $T$ policy decisions.
Here, $T$ counts policy decisions, each producing an $H$-step action chunk.
For the analysis, $s$ denotes an augmented Markov state. 
We write the student policy as $\pi_\phi(s)$.
All policies share the same dynamics, initial-state distribution, execution rule, and task reward.
For any policy $\pi$, let $J(\pi)=\mathbb{E}_{\pi}[\sum_{t=1}^{T}r_t]$ denote its expected cumulative task reward with no discounting. 
$\pi^*$ is the optimal policy by maximizing $J$.
Let $d_t^{\pi_\phi}$ denote the distribution of $s_t$ induced by autonomous execution of the student policy $\pi_\phi$. We define its time-averaged state distribution as $d^{\pi_\phi}=T^{-1}\sum_{t=1}^{T}d_t^{\pi_\phi}$.

Using the standardized teacher target $\mathbf{A}^{T}$ and student prediction $\mathbf{A}^{S}=\pi_\phi(s)$, define the BC loss's conditional mean and population average:
\[
\begin{aligned}
    \ell_{\mathrm{BC}}(s,\phi)
        &=\mathbb{E}_{\mathbf{A}^{T}\mid s}
          \bigl[\|\mathbf{A}^{S}-\mathbf{A}^{T}\|_2\bigr],\\
    \epsilon_{\mathrm{fit}}
        &=\mathbb{E}_{s\sim q}[\ell_{\mathrm{BC}}(s,\phi)].
\end{aligned}
\]
Both chunks are flattened for the norm, as in the BC objectives below; the conditional expectation averages over teacher targets at a fixed state and task instruction $\ell$.
Let $Q_t^{\mathrm{teacher}}(s,\mathbf{A})$ denote the expected remaining cumulative task reward after choosing chunk $\mathbf{A}$ and subsequently following $\pi_{\mathrm{teacher}}$ through the same encoding and execution procedure.

\begin{theorem}[Policy Adaptation Performance Bound]
\label{thm:unified_adaptation}
Suppose $\pi_\phi$ is deterministic, $0\le\ell_{\mathrm{BC}}(s,\phi)\le B$ on the supports of $q$ and $d^{\pi_\phi}$, and $Q_t^{\mathrm{teacher}}(s,\cdot)$ is uniformly $L$-Lipschitz in the chunk norm for the relevant student and teacher actions.
Then, the student's reward shortfall relative to the optimum is bounded by
\begin{equation}
    J(\pi^*)-J(\pi_\phi)\le LT\bigl[
    \underbrace{\mathstrut\epsilon_{\mathrm{fit}}}_{\substack{\text{fitting}\\\text{error}}}
    +\underbrace{\mathstrut B\,\mathrm{TV}(d^{\pi_\phi},q)}_{\substack{\text{state distribution}\\\text{mismatch}}}
    \bigr]+\underbrace{\mathstrut\Delta_{\mathrm{teacher}}}_{\substack{\text{reference}\\\text{quality}}}.
\label{eq:unified_adaptation_bound}
\end{equation}
\end{theorem}

The first term measures the population action-fitting error under the supervision distribution $q$.
The second accounts for the difference between training states and autonomous deployment states, where $\mathrm{TV}(p,q)=\sup_E|p(E)-q(E)|$ is total variation distance and $B$ bounds the nonnegative action loss.
The factor $L$ measures how strongly action deviations can affect the teacher's remaining cumulative task reward.
Finally, $\Delta_{\mathrm{teacher}}=J(\pi^*)-J(\pi_{\mathrm{teacher}})\ge0$ measures the teacher's suboptimality as the reference: a smaller value means a better reference.
A detailed proof is presented in Appendix~\ref{app:unified_adaptation}.

This theorem explains the three stages.
Offline distillation uses targets from $\pi_{\mathrm{teacher}}$ and teacher rollouts as $q$, transferring its behavior by reducing action-fitting error.
OPD retains the teacher reference but adds labels at student-visited states, targeting the distribution mismatch\cite{ross2011reduction}.
IPR therefore addresses reference quality last: human corrections add alternative behaviors, and IQL-style value weighting favors recorded actions with higher estimated returns~\cite{kostrikov2022iql}.
Using only a subset of these stages makes it difficult to reduce all three error terms simultaneously.
The remainder of this section follows these three stages.
We first describe offline distillation and OPD with masked-action supervision, then detail IPR.

\subsection{Offline Distillation from a Teacher Policy}

Directly deploying a policy with random weight can be unsafe on a physical humanoid.
We therefore begin with offline distillation and do not use the student to collect online experience at this stage~\cite{ning2026proxypolicysteering}.
The goal is to transfer the teacher's behavior patterns to the student under the teacher-induced state distribution, providing a safer initialization for the subsequent OPD and RL stages.

We rollout the frozen teacher $\pi_{\mathrm{teacher}}$ in the target environment and collect observation--action pairs into the teacher-generated dataset
$
    \mathcal{D}_{\mathrm{offline}}
        = \left\{\left(\mathbf{o}_t,\mathbf{u}^T_t\right)\right\},
$
where $\mathbf{u}^T_t$ is the absolute action executed by the teacher after observing $\mathbf{o}_t$.
We store the absolute commands without delta encoding.
After data collection, we train the student policy using the BC loss.
We sample an $H$-step window of stored absolute teacher actions following $\mathbf{o}_t$, denoted by $\mathbf{U}^T_t=(\mathbf{u}^T_t,\ldots,\mathbf{u}^T_{t+H-1})$.
At trajectory boundaries, absolute-action windows are padded by repeating the last recorded command before delta encoding.
We encode each action in this window relative to the same recorded command $\mathbf{u}_{t-1}$ using Eq.~\eqref{eq:delta_action}.
The encoded action is then standardized to obtain the learning target $\mathbf{A}^{T}_t$.

We implement the student as a compact Transformer encoder--decoder with a pretrained DINOv3 ViT-S/16 \cite{simeoni2025dinov3} visual backbone.
For each camera, adaptive average pooling reduces the DINOv3 patch features to a $4\times4$ grid, yielding 16 visual tokens with 384 features each.
The visual tokens and proprioceptive state are linearly projected into a shared 512-dimensional embedding space, with the state represented by a single token.
The Transformer encoder fuses these representations, and the decoder uses $H$ learned action queries followed by a linear regression head to predict the action chunk.
The network directly outputs all $H$ chunk-wise delta actions in one forward pass.

For a training pair $(\mathbf{o}_t,\mathbf{A}^{T}_t)$, the BC loss compares the student's prediction with the teacher target:
\begin{equation}
    \ell_{\mathrm{BC}}
    = \left\|\mathbf{A}^{S}_t-\mathbf{A}^{T}_t\right\|_2,
    \label{eq:offline_bc_loss}
\end{equation}
where each chunk is flattened into a single vector for the norm.
The offline objective $\mathcal{L}_{\mathrm{BC}}$ averages $\ell_{\mathrm{BC}}$ over training pairs sampled from $\mathcal{D}_{\mathrm{offline}}$.

\subsection{OPD from Pretrained Teacher Policy}

Low imitation error on teacher demonstrations does not guarantee reliable task execution by the student.
During deployment, small action errors can drive the student into states absent from the offline data, allowing deviations to compound into task failure.
We address this mismatch with OPD: the student collects trajectories with teacher assistance, and the frozen teacher relabels the collected states.
OPD provides a dense corrective target at every student visited state.

\begin{figure}[!hb]
    \centering
    \includegraphics[width=\linewidth]{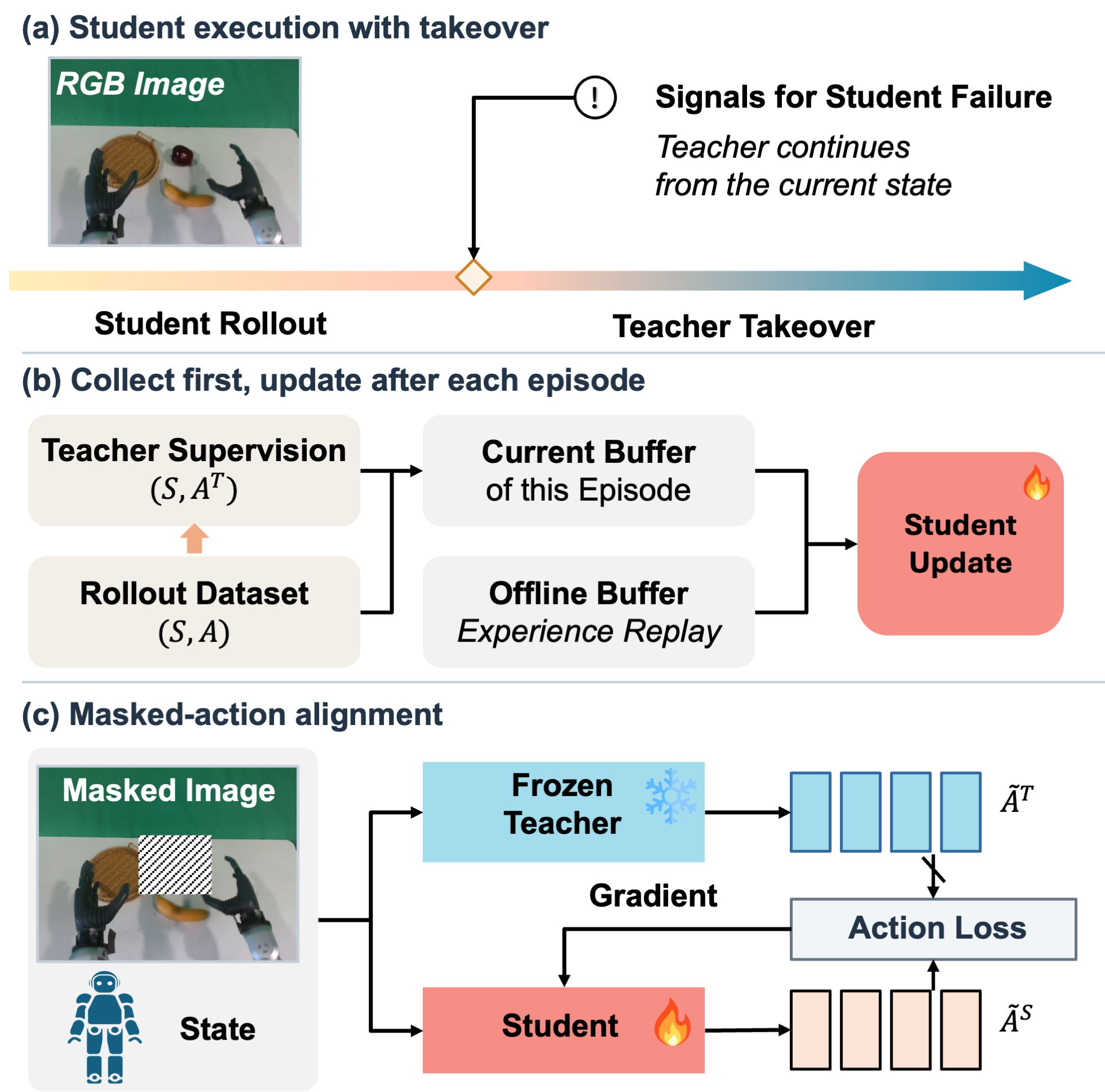}
    \caption{Overview of OPD: (a) the teacher takes over after a human signals student failure; (b) student updates follow trajectory collection, combining teacher supervision on the current rollout with replay of the fixed offline teacher dataset; and (c) masked-action matching transfers teacher predictions under the same masked observation.}
    \label{fig:opd_framework}
\end{figure}

Bridging OPD from discrete language modeling~\cite{agarwal2024onpolicy} to continuous whole-body manipulation requires addressing three embodied physical constraints: horizon reachability, multi-configuration retention, and physical interaction scarcity.
We introduce three principled mechanisms accordingly (Fig.~\ref{fig:opd_framework}):
\begin{enumerate}[leftmargin=*]
    \item \textbf{Teacher-Takeover Horizon Completion:} Whenever an early-stage student deviates toward failure, the frozen teacher seamlessly assumes control to advance the rollout through downstream phases, preventing supervision from collapsing onto initial reaching segments.
    \item \textbf{Anchor-Distribution Replay:} Interleaving newly relabeled student states with the fixed offline anchor buffer $\mathcal{D}_{\mathrm{offline}}$ preserves spatial generalization across previously mastered object layouts during sequential online updates.
    \item \textbf{Counterfactual Masked-Action Alignment:} Matching student and teacher action responses under identical structured visual occlusions transfers task-causal visual attention purely through the black-box action interface, doubling supervision without requiring additional rollouts.
\end{enumerate}

\subsubsection{Asynchronous Rollout Collection and Teacher Takeover}
Querying the teacher policy and updating the student at every robot action incurs substantial computation and control latency, which can reduce real-world task success.
Therefore, we update the policy after a complete trajectory collection has finished.
At iteration $j$, the student $\pi_{\phi_j}$ begins the rollout with its parameters fixed throughout collection.
A human operator monitors execution and signals whether the student has failed.
Upon this signal, the frozen teacher policy takes over, infers actions from current observations, and continues the task for a more complete trajectory.

We record observations and executed absolute actions throughout the rollout.
After collection, we query the teacher on states visited under student control to obtain imitation targets and retain recorded teacher actions as demonstration targets.
All target chunks start at decision step $t$ and are conditioned on the recorded observation $\mathbf{o}_t$.
Each chunk uses the recorded $\mathbf{u}_{t-1}$ as the fixed reference for all its actions and is standardized as in offline BC.
Policy updates and masked teacher queries occur after collection, keeping them outside the robot's control loop.

\subsubsection{Counterfactual Masked-Action Alignment}Real-world trajectories are scarce and costly to collect, motivating us to extract more supervision from each recorded observation.
We introduce counterfactual masked-action alignment, inspired by counterfactual reasoning, to probe the teacher's dependence on local visual information through controlled image occlusion.
For each current-rollout state--action pair, we sample the number of rectangular masks uniformly from $\{1,2,3\}$ and apply them to the RGB inputs $\mathbf{I}_t$ in $\mathbf{o}_t$, obtaining $\widetilde{\mathbf{o}}_t$ while keeping proprioception and the teacher's task instruction unchanged.
Each rectangle has a width-to-height ratio sampled uniformly from $[0.5,2.0]$ and an area between $10\%$ and $30\%$ of the total image area.
This intervention on the visual input poses a counterfactual question: what action would the teacher predict if particular visual evidence were unavailable while the remaining inputs stayed fixed?
The frozen teacher supplies a masked-observation target $\widetilde{\mathbf{A}}^{T}_t$, and the student predicts $\widetilde{\mathbf{A}}^{S}_t=\pi_\phi(\widetilde{\mathbf{o}}_t)$.
We generate the masked observation and its teacher target only once per original state--action pair and reuse them across optimization epochs.
Combined with BC on clean observations, masked-action matching encourages the student to reproduce the teacher's sensitivity to the removal of visual evidence.
This supplies one additional teacher-supervised example per recorded pair without further robot interaction, aiming to improve sample efficiency.
The alignment loss consists of masked-action matching:
\begin{equation}
    \ell_{\mathrm{align}}
    = \left\|\widetilde{\mathbf{A}}^{S}_t
    -\widetilde{\mathbf{A}}^{T}_t\right\|_2.
    \label{eq:opd_mask_loss}
\end{equation}
Clean and masked teacher chunks share the same recorded command $\mathbf{u}_{t-1}$ as the fixed reference for all their actions.
Both use the same delta encoding and standardization as the offline BC targets.

\subsubsection{Composite OPD Objective}

The OPD objective combines supervision from the current rollout and the fixed offline teacher dataset:
\begin{equation}
    \mathcal{L}_{\mathrm{OPD}}
    = (1-\rho)\mathcal{L}_{\mathrm{cur}}
    + \rho\mathcal{L}_{\mathrm{rep}},
    \label{eq:opd_total_loss}
\end{equation}
where $\rho=0.3$ controls the contribution of offline replay.
The current-rollout loss $\mathcal{L}_{\mathrm{cur}}$ combines BC and masked-action alignment to learn from newly collected states.
The replay loss $\mathcal{L}_{\mathrm{rep}}$ uses BC on the initial offline teacher data to retain behavior learned during offline distillation.
Each update draws batches $\mathcal{B}_{\mathrm{cur}}$ from the current rollout and $\mathcal{B}_{\mathrm{rep}}$ from $\mathcal{D}_{\mathrm{offline}}$.
The dataset $\mathcal{D}_{\mathrm{offline}}$ remains fixed throughout OPD and contains only the teacher rollouts collected for offline warm-up.
Replaying these trajectories helps preserve the student's initial behavior as it adapts to newly visited states.
The current-rollout and replay losses are computed over their batches:
\begin{equation}
\begin{aligned}
    &\mathcal{L}_{\mathrm{cur}}
        = \mathbb{E}_{\mathcal{B}_{\mathrm{cur}}}\!\left[
        \alpha\ell_{\mathrm{BC}}
        + (1-\alpha)\ell_{\mathrm{align}}\right], \\
    &\mathcal{L}_{\mathrm{rep}}
        = \mathbb{E}_{\mathcal{B}_{\mathrm{rep}}}\!\left[
        \ell_{\mathrm{BC}}\right].
\end{aligned}
\label{eq:opd_batch_losses}
\end{equation}
Here, $\alpha=0.7$ balances BC and alignment, and each expectation denotes an empirical batch average.
During this stage, teacher targets remain fixed, and gradients update only the student parameters.
Masked observations are used only for training; the robot executes actions predicted from clean observations.
After completing the updates, the new student begins the next rollout.

\subsection{Intervention-guided Policy Refinement}

After OPD training, the student policy can achieve a success rate close to that of the teacher.
However, the teacher itself may have a low success rate in OOD scenarios.
To address this limitation, we introduce IPR, a policy refinement module that combines whole-body human intervention with IQL.
During deployment, a human operator provides corrective actions at failure-prone or recovery-critical states, thereby supplying supervision in regions that are poorly covered by the teacher-generated data.
IQL then uses the collected task rewards and intervention trajectories to favor actions associated with successful outcomes.

\subsubsection{Whole-Body Human Intervention}

We use the same PICO~4 interface as in the teleoperation system described above.
Directly switching from autonomous control to absolute-pose teleoperation would require the operator and robot poses to be closely aligned at the handover instant, which is cumbersome for whole-body control involving multiple anchors.
Some approaches introduce a transition phase that interpolates from the robot's current pose to the teleoperation target before handing over control \cite{xiao2026rove}; this avoids abrupt motion changes but reduces data quality by adding alignment segments to the dataset.
We therefore introduce a delta-pose intervention scheme that preserves the current robot configuration as the reference and transfers only the operator's motion relative to the beginning of the intervention.

During an autonomous rollout, the operator monitors the robot and requests an intervention when the policy produces an incorrect or failure-prone behavior.
Once the request is received, policy-predicted actions are no longer executed.
The corresponding policy actions and observations during this handover period are also excluded from the replay buffer.
The operator then coarsely aligns their body with the robot and begins the intervention at time $t_0$.
For each anchor $i\in\mathcal{K}$, let $\mathbf{q}^{H,i}_t$ denote the human pose after retargeting to the robot's task space.
Let $\mathbf{q}^{R,i}_t$ denote the corresponding robot pose, with both pose variables defined in $\mathrm{SE}(3)$.
For each subsequent intervention step, we compute
\begin{equation}
\begin{aligned}
    &\Delta\mathbf{q}^{H,i}_t
        = \mathbf{q}^{H,i}_t \ominus \mathbf{q}^{H,i}_{t_0}, \\
    &\mathbf{q}^{R,i,*}_t
        = \mathbf{q}^{R,i}_{t_0}\oplus
        \Delta\mathbf{q}^{H,i}_t,
    \ i\in\mathcal{K},
\end{aligned}
\label{eq:intervention_delta_retarget}
\end{equation}
where $\mathbf{q}^{R,i,*}_t$ is the commanded robot pose, and $\ominus$ and $\oplus$ denote relative-pose extraction and pose composition, respectively.
This formulation only requires coarse initial alignment because mismatches between the absolute human and robot poses are not directly transferred.

After teleoperation begins, the human corrective actions actually executed by the robot are stored in $\mathcal{B}$ together with their transitions; the suspended policy predictions are never used as executed actions.

When the operator releases control at time $t_{\mathrm{exit}}$, the current observation $\mathbf{o}_{t_{\mathrm{exit}}}$ is passed back to the policy to infer a new action chunk.
With conventional absolute-action prediction, the policy can pull the robot toward configurations represented in its training distribution when human control ends.
A mismatch between these configurations and the final human-controlled configuration can produce a large command jump at handover.
Such jumps can damage hardware and degrade dataset quality by introducing abrupt transitions into the recorded trajectories.
With our chunk-wise delta representation, each continuous command is reconstructed by adding its predicted delta to the last human command $\mathbf{u}_{t_{\mathrm{exit}}-1}$.
For states within the training distribution, the first predicted continuous delta is typically small, limiting the command change and supporting a smooth handover from human to policy control.
We demonstrate the effectiveness of this intervention scheme in Section~\ref{sec:eval_whole_body_intervention} through recorded command trajectories around handover.

\subsubsection{Policy Updates in IPR}
IPR combines DAgger-style data aggregation, intervention-aware credit assignment, and Q-chunk value estimation within IQL.
This combination is motivated by two constraints of real-world humanoid learning.
First, on-policy methods require large amounts of newly collected experience after each policy update, making them prohibitively sample-inefficient on a physical robot.
Second, conventional off-policy actor--critic methods improve the actor by maximizing its learned Q-function.
In high-dimensional whole-body action space, a limited interaction dataset covers only a small fraction of the possible actions, so direct Q maximization can exploit erroneous values extrapolated to unsupported action combinations.
IQL avoids this maximization: it fits values using only actions present in the replay buffer and extracts the policy through advantage-weighted regression (AWR) on those actions ~\cite{kostrikov2022iql}.
It therefore retains replay-based sample efficiency while reducing OOD action queries during policy improvement.

\paragraph{Q-chunk value learning}
A one-step critic evaluates an individual action even though our policy produces coordinated whole-body action chunks.
We therefore extend IQL with Q-chunking to evaluate complete sequences of coordinated base, body-anchor, and hand commands.
Chunk-level backups propagate delayed rewards across multiple transitions in one update, reducing the number of bootstrapping stages needed for long-horizon credit assignment.
Conditioning the critic on the executed action sequence also avoids the intermediate-action mismatch that biases naive off-policy multi-step returns.
Together, these properties motivate Q-chunking for more stable and sample-efficient value learning under our limited real-world interaction budget~\cite{li2025qchunking}.

For each human takeover, we assign a failure penalty of $-10$ to the final executed action of the preceding autonomous rollout segment.
At the end of each episode, we assign a terminal reward of $+10$ for success and $-10$ for failure.

We sample action chunks of length $H$ from replay trajectories.
As in BC, absolute-action windows near the trajectory end are padded to length $H$ by repeating the last recorded command before delta encoding.
For a chunk conditioned on $\mathbf{o}_t$, the sampled absolute actions begin at $t$.
We encode all actions relative to the same recorded command $\mathbf{u}_{t-1}$ using Eq.~\eqref{eq:delta_action} and standardize the resulting chunk as in BC.

Let $\mathbf{o}_{t'}$ be the next observation after the last recorded transition in the sampled chunk.
The discounted chunk reward is
$
    R_t = \sum_{k=t}^{t'-1}\gamma^{k-t}r_k.
\label{eq:ipr_truncated_chunk}
$
Let $d^{\mathrm{chunk}}_t$ indicate whether the sampled segment ends at a true terminal transition.

\paragraph{Implicit Q-learning}
We use IQL~\cite{kostrikov2022iql} to learn a state-value function $V_\psi$ and an ensemble of action-chunk critics $\{Q_{\theta_i}\}_{i=1}^{N}$.
IQL forms bootstrap targets from $V_\psi$, avoiding explicit maximization of $Q$ over action chunks that may lie outside the replay distribution.
To reduce overestimation, we use the minimum target-critic prediction across the ensemble, $\bar Q_t=\min_{i=1,\ldots,N}Q_{\bar\theta_i}(\mathbf{o}_t,\mathbf{A}_t)$, where $N$ is the ensemble size and $\bar\theta_i$ denotes the target-network parameters.

We fit $V_\psi$ to replay action values via expectile regression:
\begin{equation}
\begin{aligned}
    &\mathcal{L}_{V}(\psi)
        = \mathbb{E}_{\mathcal{B}}
        \left[L_{\tau}\!\left(\bar Q_t-V_\psi(\mathbf{o}_t)\right)\right], \\
    &L_{\tau}(u)
        = \left|\tau-\mathbb{I}[u<0]\right|u^2.
\end{aligned}
\label{eq:ipr_value_loss}
\end{equation}
Here, $\mathbb{E}_{\mathcal{B}}$ averages over sampled replay chunks.
Choosing $\tau\in(0.5,1)$ gives greater weight to positive residuals, directing $V_\psi$ toward the higher values among action chunks represented in replay.
In this work, we set $\tau = 0.7$.
Each critic then learns from the observed chunk return and the value of the next recorded observation:
\begin{equation}
\begin{aligned}
    &y_t
        = R_t
        + \gamma^{t'-t}\bigl(1-d^{\mathrm{chunk}}_t\bigr)
          V_\psi(\mathbf{o}_{t'}), \\
    &\mathcal{L}_{Q}(\theta_i)
        = \mathbb{E}_{\mathcal{B}}
        \!\left[
        \left(Q_{\theta_i}(\mathbf{o}_t,\mathbf{A}_t)
        - \operatorname{sg}(y_t)\right)^2\right],
\end{aligned}
\label{eq:ipr_q_loss}
\end{equation}
The operator $\operatorname{sg}$ holds the bootstrap target fixed during each critic update.
A true task termination, whether success or failure, sets $d^{\mathrm{chunk}}_t=1$ and removes the bootstrap term.
A human takeover only changes the controller and does not itself terminate the episode: the intervention penalty changes the stored reward but leaves bootstrapping enabled.
Timeouts and finite recording boundaries likewise retain $d^{\mathrm{chunk}}_t=0$ and bootstrap from the next recorded observation $\mathbf{o}_{t'}$.

Finally, we improve the student policy through advantage-weighted regression on action chunks from the replay buffer.
The advantage $\bar Q_t-V_\psi(\mathbf{o}_t)$ measures how much better a recorded chunk is than the state's learned value baseline.
We assign higher-advantage chunks larger regression weights and clip these weights to prevent a few samples from dominating the update:
\begin{equation}
\begin{aligned}
    &w_t
        = \min\!\left(w_{\max},
        \exp\!\left(\frac{\bar Q_t-V_\psi(\mathbf{o}_t)}{\beta}\right)\right),\\
    &\mathcal{L}_{\pi}(\phi)
        = \mathbb{E}_{\mathcal{B}}\!\left[
        w_t\,\left\|\mathbf{A}^{S}_t-\mathbf{A}_t\right\|_2
        \right].
\end{aligned}
\label{eq:ipr_actor_loss}
\end{equation}
Here, the unsquared $L_2$ norm is computed by flattening the action vectors from recorded steps only, excluding padded actions from the loss.
The temperature $\beta>0$ controls how strongly the weights favor high-advantage chunks, and $w_{\max}$ caps each weight.

\subsubsection{The Complete System of RAPID}

Algorithm~\ref{alg:rapid-training} summarizes the three training stages for the compact student: offline distillation, OPD, and IPR.

\begin{algorithm}[t]
\caption{The framework of \methodname System}
\label{alg:rapid-training}
\small
\DontPrintSemicolon
\SetAlgoNoEnd
\SetKwBlock{Parallel}{in parallel}{end}
\KwIn{Pretrained Policy $\pi_{\mathrm{teacher}}$; task instruction $\ell$}
\KwOut{Task-adapted student $\pi_{\phi^\star}$}
Freeze $\pi_{\mathrm{teacher}}$ and initialize the compact student $\pi_\phi$\;

\tcp{Stage 1: Offline distillation}
Collect teacher rollouts into $\mathcal{D}_{\mathrm{offline}}$\;
Train $\pi_\phi$ on $\mathcal{D}_{\mathrm{offline}}$ with BC (Eq.~\eqref{eq:offline_bc_loss})\;

\tcp{Stage 2: On-policy distillation}
\For{each OPD rollout}{
    Collect a student trajectory with teacher takeover\;
    Query the teacher on student-visited states and retain recorded teacher actions as targets\;
    Update $\pi_\phi$ with current-rollout supervision, replay from $\mathcal{D}_{\mathrm{offline}}$, and mask alignment (Eq.~\eqref{eq:opd_total_loss})\;
}

\tcp{Stage 3: IPR}
Initialize $Q_\theta$, $V_\psi$, target critics, and replay $\mathcal{B}\gets\varnothing$\;
\Parallel{
    \tcp{Actor thread}
    \While{training continues}{
        Interact using the latest received policy, with whole-body human intervention\;
        Append executed transitions with task rewards and takeover penalties to $\mathcal{B}$\;
    }
    \tcp{Learner thread}
    \While{training continues}{
        Sample action chunks from $\mathcal{B}$ when available\;
        Update $Q_\theta$, $V_\psi$, and $\pi_\phi$ with Q-chunk IQL\;
        Publish updated policy parameters $\phi$ to the actor\;
    }
}
$\phi^\star\gets\phi$; \Return{$\pi_{\phi^\star}$}\;
\end{algorithm}

%% file: sec/experiments.tex
\section{Experiments}
Our platform for real-world experiments is the LimX Oli humanoid robot.
The robot is equipped with two BrainCo Revo 1 dexterous hands and an Intel RealSense D435i camera mounted on its head for visual observation.
For the \textit{Clean Trash} and \textit{Sort Clothes} tasks, we additionally use an MV-CB016-10UC-S wrist-mounted camera.
During the experiments, policy inference runs on a local workstation equipped with an NVIDIA GeForce RTX 4090 GPU, an Intel Core i9-14900KF CPU, and 64~GB of RAM.
We use NVIDIA GR00T N1.7 \cite{bjorck2025gr00t} as the base teacher policy for the real-world experiments.
The base teacher policy is trained on an NVIDIA H20 GPU server and subsequently deployed on the local workstation for policy inference.
All online RL algorithms in the comparative experiments are trained on the local RTX 4090 workstation, whereas HG-DAgger is trained on the H20 server.
Our experiments address three questions:
\begin{itemize}
    \item How does \methodname perform relative to \textbf{baseline methods} on real-world tasks?
    \item How does each \textbf{component} contribute to the performance of \methodname?
    \item How well does \methodname perform across \textbf{different teacher policies and tasks}?
\end{itemize}

\subsection{Real-World Comparison}
This section presents two sets of comparisons on real-world tasks.
First, we compare \methodname with the baselines in terms of success-rate improvement, training efficiency (time and compute), deployment compute, and trajectory length under the final policies.
Second, we compare our proposed whole-body intervention scheme with a conventional absolute-action scheme, focusing on command continuity at the human-to-policy handoff.

\subsubsection{Quantitative Comparison in Real-World}
\begin{table*}[t] 
	\caption{Comparison of different policies on the Pick Fruit and Hang Cup tasks.
	Trajectory lengths are reported as mean $\pm$ standard deviation (SD) across all evaluation trajectories.}
	\centering 
	\renewcommand{\arraystretch}{1.2}
	\setlength{\tabcolsep}{3pt}
	\begin{tabular}{@{}l c c c c c c c c c@{}}
		\toprule
		\multirow{3}{*}{\textbf{Method}}
		& \multicolumn{4}{c}{\textbf{\textit{Pick Fruit}}}
		& \multicolumn{4}{c}{\textbf{\textit{Hang Cup}}}
		& \multirow{3}{*}{\makecell{\textbf{Deployment}\\\textbf{Compute}\\\textbf{(GFLOPs)}}} \\
		\cmidrule(lr){2-5}\cmidrule(lr){6-9}
		& \makecell{Success\\Rate}
		& \makecell{Training\\Time (h)}
		& \makecell{Training\\Compute (PFLOPs)}
		& \makecell{Trajectory\\Length (s)}
		& \makecell{Success\\Rate}
		& \makecell{Training\\Time (h)}
		& \makecell{Training\\Compute (PFLOPs)}
		& \makecell{Trajectory\\Length (s)} & \\
		\midrule
		Base Policy & 37.04\% & --    & --     & $7.33 \pm 0.82$  & 15.00\% & --   & --     & $\mathbf{10.74 \pm 0.73}$ & 771.75 \\
		HG-DAgger \cite{kelly2019hg}  & 77.78\% & 6.45  & 203.59 & $11.48 \pm 2.15$ & 35.00\% & 4.97 & 367.68 & $12.59 \pm 1.35$ & 771.75 \\
		DSRL \cite{wagenmaker2025steering}     & 55.56\% & 0.97 & 6.66   & $7.33 \pm 1.55$  & 25.00\% & 0.80 & 5.82   & $11.28 \pm 0.84$ & 733.41 \\
		RL Token \cite{xu2026rl}   & 48.15\% & 1.12 & 21.66  & $\mathbf{5.85 \pm 0.30}$  & 45.00\% & 0.72 & 13.98  & $10.77 \pm 0.85$ & 389.54 \\
		\midrule
		\textbf{Ours} & \textbf{88.89\%} & 1.00 & 15.49 & $13.00 \pm 1.81$ & \textbf{80.00\%} & 0.95 & 13.86 & $18.91 \pm 0.97$ & \textbf{10.20} \\
		\bottomrule
	\end{tabular}
	\label{tab:success_rate}
\end{table*}

\suppressfloats[t]
\begin{figure}[t]
    \centering
    \includegraphics[width=\columnwidth]{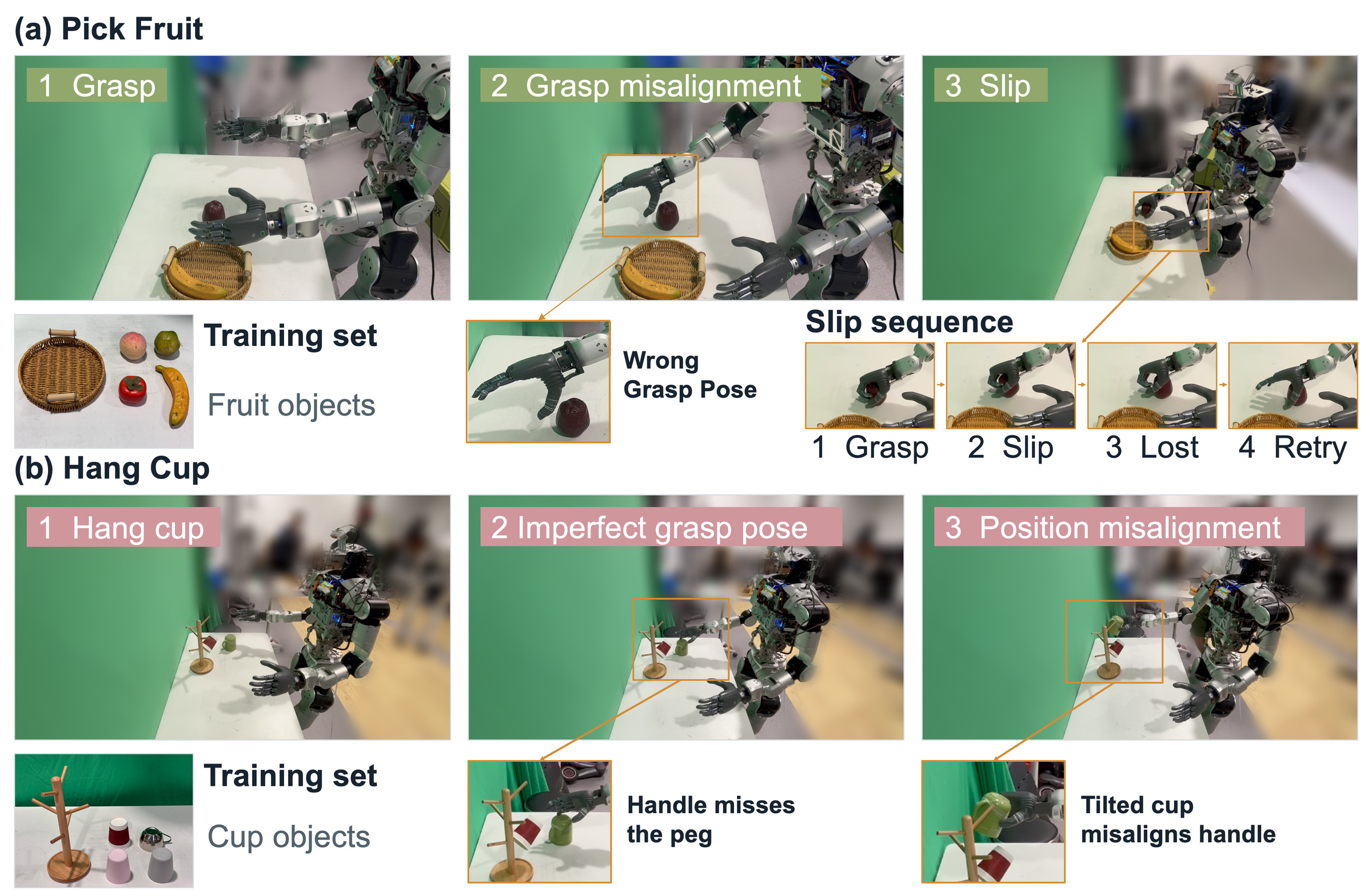}
    \caption{Task settings, training objects, and representative deployment failures for (a) Pick Fruit and (b) Hang Cup.
    Pick Fruit failures include grasp misalignment and object slipping; Hang Cup failures include imperfect grasp poses and handle--peg misalignment.}
    \label{fig:real_world_scene}
\end{figure}

\begin{table}[t]
    \caption{Training data for each real-world task, reported as [base-policy human demonstrations $|$ post-training interactions].
    Pair counts denote recorded observation--action pairs $(\mathbf{o}_t,\mathbf{u}_t)$.}
    \label{tab:training_data}
    \centering
    \small
    \renewcommand{\arraystretch}{1.2}
    \setlength{\tabcolsep}{3pt}
    \begin{tabular*}{\columnwidth}{@{\extracolsep{\fill}} l C{2em}@{\,$|$\,}C{2em} C{2.8em}@{\,$|$\,}C{2.8em} C{2.8em}@{\,$|$\,}C{2.8em} @{}}
        \toprule
        \textbf{Task}
        & \multicolumn{2}{c}{\textbf{\# of trajs.}}
        & \multicolumn{2}{c}{\textbf{\# of pairs}}
        & \multicolumn{2}{c}{\textbf{Time (h)}} \\
        \midrule
        \texttt{Pick Fruit}  & 260 & 120 & 123.6k & 42.5k & 1.145 & 0.393 \\
        \texttt{Hang Cup}    & 107 & 100 & 52.6k & 46.1k & 0.487 & 0.427 \\
        \texttt{Clean Trash} & 52  & 100  & 13.8k & 67.0k & 0.128 & 0.620 \\
        \texttt{Sort Clothes} & 60 & 81 & 18.4k & 45.2k & 0.170 & 0.419 \\
        \bottomrule
    \end{tabular*}
\end{table}

We design the experiments to simulate OOD scenarios that a pretrained policy may encounter during real-world deployment.
For each task, we collect human demonstrations $\mathcal{D}_{\mathrm{base}}$ that differ from the test distribution and use them to fine-tune GR00T N1.7~\cite{bjorck2025gr00t} into the base teacher policy.
Table~\ref{tab:training_data} distinguishes these demonstrations from post-training interactions, which include the initial teacher rollouts used to populate $\mathcal{D}_{\mathrm{offline}}$ for student warm-up and the subsequent OPD and IPR rollouts.

The \textit{Pick Fruit} teacher is fine-tuned on 260 human demonstration trajectories involving several types of fruit, including apples, bananas, and peaches.
Fruit positions vary across the demonstrations.
During evaluation, we fix the target object to an apple and place it at nine predefined locations.
We evaluate success rate over three trials at each location, for a total of 27 trials.
By holding the object category constant, this protocol isolates spatial variation and evaluates how quickly \methodname adapts across object positions.
The \textit{Hang Cup} teacher is fine-tuned on 107 human demonstration trajectories involving other types of cups.
During evaluation, the robot is required to hang a green plastic cup on the cup rack.
We evaluate success rate over a total of 20 trials.
This change in cup type evaluates \methodname's ability to generalize across cup categories.
Fig.~\ref{fig:real_world_scene} illustrates the variation in the demonstration data and representative deployment failures for both tasks.

We compare \methodname with HG-DAgger~\cite{kelly2019hg}, DSRL~\cite{wagenmaker2025steering}, and RL Token~\cite{xu2026rl}.
In our experiments, HG-DAgger, RL Token, and \methodname use human corrections through the same intervention interface, whereas DSRL is trained without human-intervention data.

Table~\ref{tab:success_rate} reports the final evaluation results.
On \textit{Pick Fruit}, \methodname achieves an 88.89\% success rate, exceeding the strongest baseline, HG-DAgger (77.78\%), by 11.11 percentage points.
On \textit{Hang Cup}, it achieves 80.00\%, outperforming the strongest baseline, RL Token (45.00\%), by 35.00 percentage points.
Whereas \textit{Pick Fruit} tests adaptation across object positions, \textit{Hang Cup} requires precise handle--peg alignment and fine corrective motions using RGB observations without explicit depth input.

\begin{figure}[!t]
    \centering
    \includegraphics[width=\columnwidth]{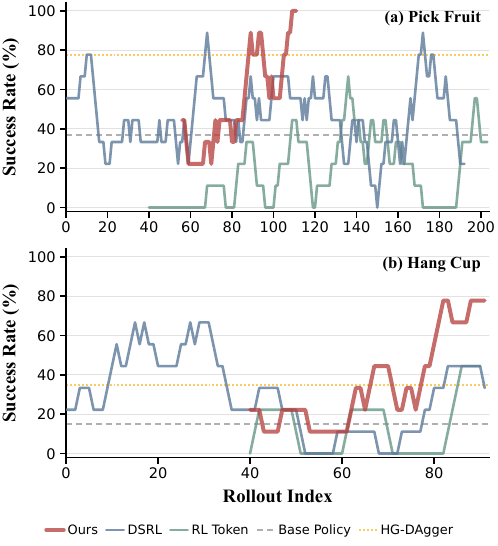}
    \caption{Autonomous success rates during online training on (a) \textit{Pick Fruit} and (b) \textit{Hang Cup}.
    Each point averages a 9-rollout window.
    A rollout counts as successful only when completed without human intervention.
    The horizontal lines show the Base Policy and HG-DAgger evaluation success rates from Table~\ref{tab:success_rate}.
    Extended DSRL and RL Token training beyond the common budget is shown only for \textit{Pick Fruit}.}
    \label{fig:success_rate_curve}
\end{figure}

Fig.~\ref{fig:success_rate_curve} complements these final evaluations by showing autonomous success rates during online training.
\methodname uses 120 interaction rollouts on \textit{Pick Fruit} and 100 on \textit{Hang Cup}, including offline warm-up, OPD, and IPR.
On \textit{Pick Fruit}, we extend DSRL and RL Token training to approximately 200 rollouts (about $1.7\times$ the \methodname budget) to assess further improvement; their results in Table~\ref{tab:success_rate} still use the common 120-rollout budget.
RL Token's 40 teacher warm-up rollouts count toward the interaction budget but are excluded from the plotted windows.
The \methodname curves show only IPR, beginning after warm-up and OPD at rollout index 56 for \textit{Pick Fruit} and 40 for \textit{Hang Cup}.
On \textit{Pick Fruit}, DSRL peaks at 88.89\% but fluctuates substantially, while RL Token peaks at 66.67\% without sustaining this improvement even under the extended budget.
\methodname reaches 100\% autonomous success in the final windows within its 120-rollout budget.
On \textit{Hang Cup}, DSRL reaches 66.67\% early in training but declines to 33.33\% in the final window; RL Token remains low for most of training and reaches 44.44\% near the end.
In contrast, \methodname improves during later IPR updates to 77.78\% in the final window.
These training curves show improved autonomous completion late in IPR on both tasks; the separate final-policy evaluations yield 88.89\% and 80.00\%, respectively (Table~\ref{tab:success_rate}).

Moreover, behavioral inspection explains execution timing ($13.00 \pm 1.81$~s on \textit{Pick Fruit} and $18.91 \pm 0.97$~s on \textit{Hang Cup} vs.\ $5.85$--$12.59$~s for baselines whose averages include rapid open-loop misses): \methodname autonomously executes closed-loop re-grasping upon incipient fruit slip and deliberately decelerates during terminal handle--peg insertion to achieve sub-centimeter RGB alignment.

\begin{table}[t]
    \caption{Success rates at different stages. Online BC and IPR start from the same post-OPD policy.}
    \label{tab:training_progress}
    \centering
    \footnotesize
    \setlength{\tabcolsep}{2pt}
    \begin{tabular*}{\columnwidth}{@{\extracolsep{\fill}} l c c c c c @{}}
        \toprule
        \textbf{Task} & \textbf{Base Policy} & \textbf{Warm-up} & \textbf{OPD} & \textbf{Online BC} & \textbf{IPR} \\
        \midrule
        \textit{Pick Fruit} & 37.04\% & 25.93\% & 59.26\% & 66.67\% & 88.89\% \\
        \textit{Hang Cup} & 15.00\% & 15.00\% & 25.00\% & 65.00\% & 80.00\% \\
        \bottomrule
    \end{tabular*}
\end{table}

Table~\ref{tab:training_progress} includes an Online BC baseline, whose comparison with IPR is presented in Section~\ref{sec:ipr_online_bc}.
It also shows how the success rates of \methodname evolve across the warm-up, OPD, and IPR stages.
After offline warm-up, the student achieves success rates of 25.93\% on \textit{Pick Fruit} and 15.00\% on \textit{Hang Cup}, suggesting that it has acquired basic task behaviors.
OPD further adapts the student to the states encountered during online interaction, raising these rates to 59.26\% and 25.00\%, respectively.
Online RL then refines the policy to achieve higher success rates of 88.89\% and 80.00\%, respectively.

\subsubsection{Efficiency Comparison}

We compare the final success-rate improvement achieved by different methods per unit of training time and compute using the results in Table~\ref{tab:success_rate}.
Throughout the paper, training time covers the full post-training process, excluding base-policy fine-tuning and separate success-rate evaluation.
For \methodname, this comprises offline student warm-up, OPD, and IPR, including all rollout collection, environment resets, clean and masked teacher queries, human interventions, data loading, and model updates.
The same accounting convention applies to all compared methods.

\methodname achieves the largest success-rate improvement within approximately one hour on both tasks: 51.85 percentage points in 1.003~h on \textit{Pick Fruit} and 65.00 percentage points in 0.95~h on \textit{Hang Cup}.
These results correspond to 51.70 and 68.42 percentage points per hour, respectively, which are $2.70\times$ and $1.64\times$ the rates of the most time-efficient baseline on each task (DSRL on \textit{Pick Fruit} and RL Token on \textit{Hang Cup}).
HG-DAgger requires $6.43\times$ and $5.23\times$ as much training time as \methodname on the two tasks, respectively, while producing smaller absolute gains.
The same trend holds after normalizing by training compute: \methodname obtains 3.35 and 4.69 percentage points per PFLOP, which are $1.20\times$ and $2.19\times$ the best baseline rates, respectively.
Thus, under this evaluation protocol, \methodname offers the strongest trade-off between success-rate improvement, training time, and compute.

\begin{figure}[!htb]
    \centering
    \includegraphics[width=\linewidth]{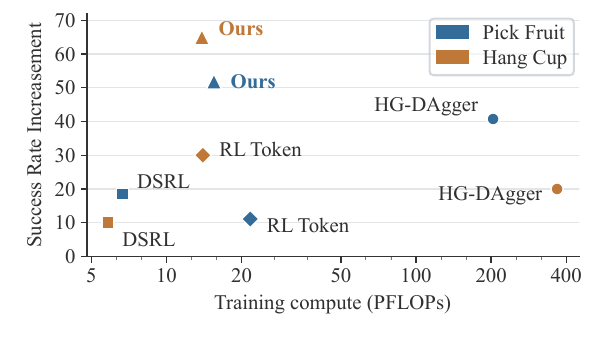}
    \caption{Training efficiency comparison across methods in terms of success-rate improvement versus training compute.}
    \label{fig:training_efficiency}
\end{figure}

For a representative visualization of training efficiency, Fig.~\ref{fig:training_efficiency} plots training compute (PFLOPs) on the horizontal axis and the final success-rate improvement over the base policy (percentage points) on the vertical axis.
The figure shows that \methodname achieves high success-rate improvements with low training compute.

We also compare deployment compute using forward-pass FLOP counts under the recorded \textit{Pick Fruit} input settings.
Compute is reported per action-chunk prediction at batch size 1, with one multiply--accumulate (MAC) counted as two floating-point operations (FLOPs).
As shown in Table~\ref{tab:success_rate}, \methodname requires 10.20~GFLOPs per prediction, reducing deployment compute by 98.68\% relative to the base policy and 97.38\% relative to RL Token.
The evaluated RL Token implementation still runs the VLA for feature extraction, whereas \methodname deploys only a compact DINOv3 encoder and ACT student.
The student runs independently of the teacher policy, substantially reducing the computational burden of repeated policy decisions during deployment.

\subsubsection{Comparison of Whole-Body Human Intervention}
\label{sec:eval_whole_body_intervention}
Our chunk-wise delta-action representation and reference-anchored takeover algorithm (Eq.~\eqref{eq:intervention_delta_retarget}) support smooth transitions into and out of human control.
In this section, we illustrate the resulting command continuity through recorded examples of the return to policy execution.

\begin{figure}[!htb]
    \centering
    \includegraphics[width=\columnwidth]{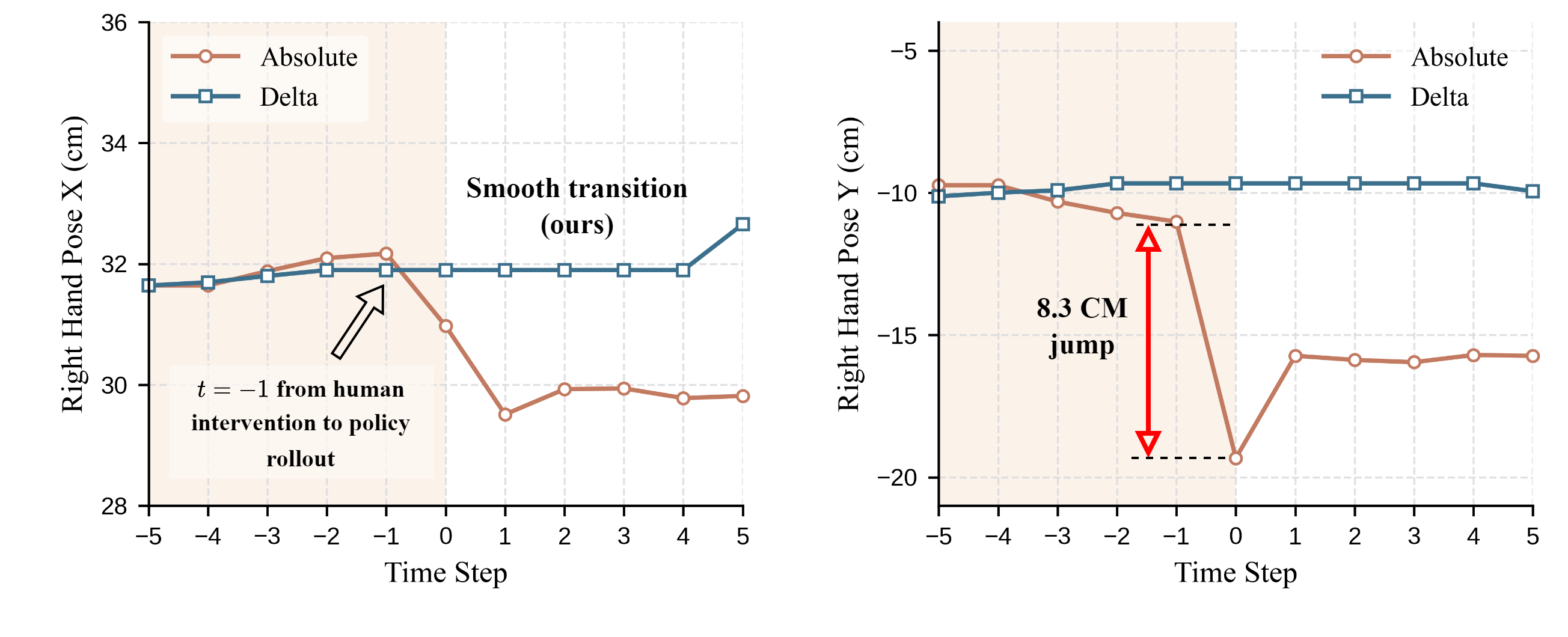}
    \caption{Illustration of right-hand commanded X (left) and Y (right) positions in two episodes.}
    \label{fig:intervention_effect}
\end{figure}

We compare our chunk-wise delta-action representation with absolute-action representation.
Detailed results are presented in Fig.~\ref{fig:intervention_effect}, which shows the trajectory when human control is released.
The figure uses local indices $-1$ and $0$ for the last human command and the first policy command, corresponding to $\mathbf{u}_{t_{\mathrm{exit}}-1}$ and $\mathbf{u}_{t_{\mathrm{exit}}}$ in our timing convention.
Between these two records, absolute-action execution produced negative jumps of $1.198$~cm in X and $8.309$~cm in Y.
With delta-action execution, we see a smooth transition between human execution and policy rollout.

Smooth transitions between human intervention and policy rollout can reduce the risk of robot hardware damage caused by abrupt command changes.
The resulting smoother trajectories also improve training data quality, helping the policy learn better behaviors.

\subsection{Ablation Studies}

We conduct the following ablation studies to address four questions.
\textbf{First}, how do teacher takeover and the allocation between offline distillation and OPD affect distillation performance?
\textbf{Second}, does mask alignment improve learning during the OPD stage?
\textbf{Third}, does initialization with OPD accelerate convergence in the subsequent online RL stage?
\textbf{Fourth}, does the RL component of IPR improve task success beyond online BC under the same interaction budget?

\subsubsection{Teacher Takeover and Offline--OPD Budget Allocation}
\label{sec:opd_takeover}

\begin{figure}[!htb]
    \centering
    \includegraphics[width=\columnwidth]{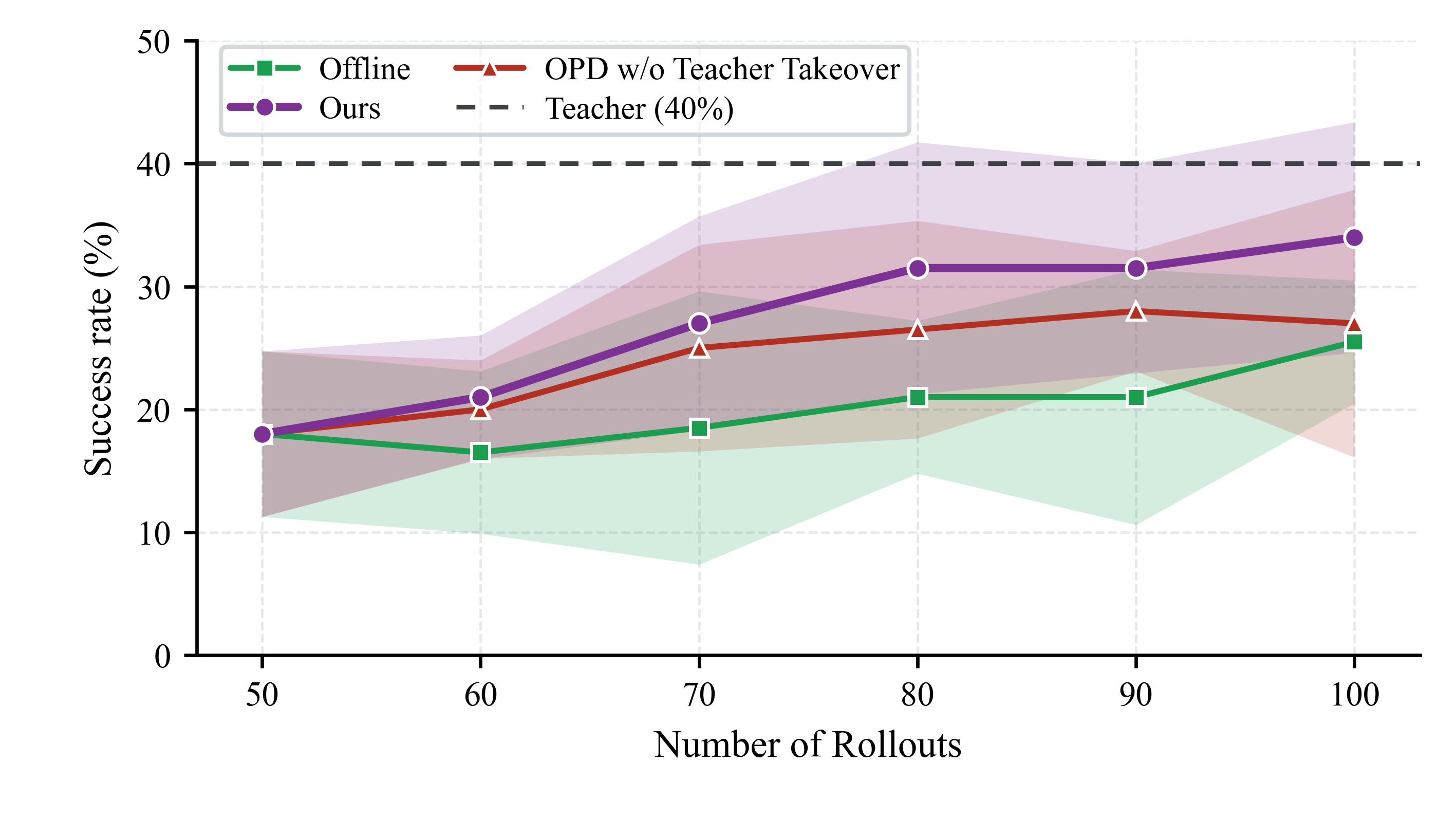}
    \caption{Ablation study on the effects of OPD with teacher takeover.
    Curves show mean success rates across four random seeds; shaded bands indicate the mean $\pm 1$ sample standard deviation across seeds.}
    \label{fig:offline_opd_compare}
\end{figure}

We compared our method with two baselines to evaluate the contribution of teacher takeover.
This experiment is conducted on the RoboTwin 2.0~\cite{chen2025robotwin} \textit{Click Bell} task.
\textit{OPD w/o Teacher Takeover} disabled teacher takeover when the student failed, while retaining OPD supervision.
\textit{Offline} trained the student through offline distillation using trajectories collected by the teacher policy.
All three methods began with offline warm-up on 50 teacher rollouts and then continued training using their respective procedures.
We evaluated each policy checkpoint over 50 episodes after every 10 training rollouts.
Each experiment was repeated across four random seeds, with results shown in Fig.~\ref{fig:offline_opd_compare}.

At 100 rollouts, \methodname achieved a mean success rate of 34.0\%, compared with 25.5\% for \textit{Offline}.
This improvement reflects OPD's use of dense teacher supervision along student-generated trajectories, rather than supervision limited to teacher-generated trajectories.
By targeting the states the student actually encounters, this supervision helps it learn more effective behaviors.

\methodname also outperformed \textit{OPD w/o Teacher Takeover}, which achieved a mean success rate of 27.0\% under the same rollout budget.
Teacher takeover broadens the training data to include trajectories the student would not be to reach independently, which helps improve policy performance.

\begin{figure}[!t]
    \centering
    \includegraphics[width=\columnwidth]{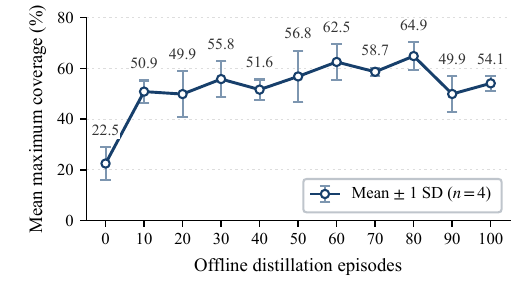}
    \caption{Effect of allocating a fixed 100-episode budget between offline distillation and OPD on \textit{Push-T}, with 20,000 student gradient updates for each allocation.
    Points show mean maximum coverage across four runs, with error bars indicating $\pm 1$ sample standard deviation across runs.}
    \label{fig:pusht_offline_opd_allocation}
\end{figure}

We additionally examine how allocating a fixed interaction budget between offline distillation and OPD affects student performance on simulated \textit{Push-T}, using the task and evaluation protocol described in Section~\ref{sec:agentic_wam_teachers}.
The student learns from a frozen GR00T N1.7-3B teacher fine-tuned on 185 Push-T demonstrations.
We fix the total budget at 100 training episodes and 20,000 student gradient updates, and evaluate all eleven offline--OPD allocations across four runs (Fig.~\ref{fig:pusht_offline_opd_allocation}).
Pure OPD without offline warm-up achieves 22.5\% mean maximum coverage, whereas offline-only distillation achieves 54.1\%.
Allocating 80 episodes to offline distillation and 20 to OPD achieves the highest mean coverage among the evaluated allocations at 64.9\%, exceeding pure OPD and offline-only distillation by 42.4 and 10.8 percentage points, respectively.
These results support combining offline initialization with OPD under fixed interaction and update budgets.

\subsubsection{Effect of OPD on Online Reinforcement Learning}

This ablation tests whether distilling the teacher policy's behavior through OPD before the final online RL stage accelerates RL convergence.
We hypothesize that OPD stage provides a stronger initialization, reducing the online interactions needed to achieve high success rates.

\begin{figure}[!htb]
    \centering
    \includegraphics[width=\columnwidth]{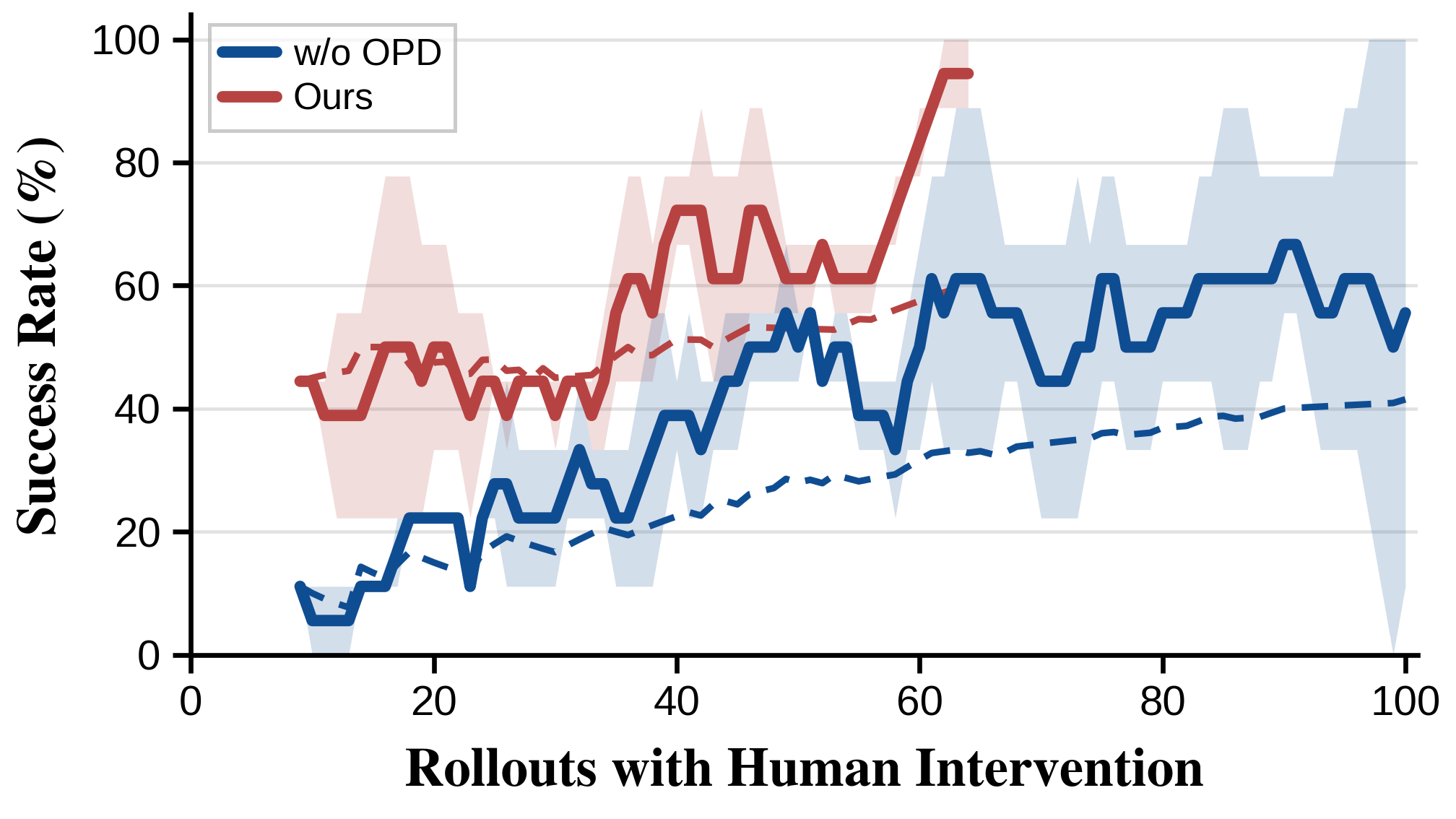}
    \caption{Effect of OPD on online RL for \textit{Pick Fruit}.
    Solid curves show the mean trailing-nine-episode success rate across two runs; shaded bands indicate their min--max range.
    Dashed curves show the mean cumulative success rate.}
    \label{fig:opd_online_rl}
\end{figure}

We conduct this experiment on the real-world \textit{Pick Fruit} task.
Under a 100-episode interaction budget, \methodname (OPD-RL) allocates 36 episodes to OPD, followed by 64 episodes of online RL.
The direct-RL baseline omits OPD and allocates all 100 episodes to the final RL stage.
We repeat each condition in two independent runs.
Fig.~\ref{fig:opd_online_rl} shows the online success rate during RL, computed within each run as the fraction of successful episodes in a trailing nine-episode window.
We average these rates pointwise across runs and shade the corresponding min--max range.

At the end of training, \methodname achieves a mean success rate of 94.44\% over the final nine-episode window, compared with 55.56\% for \textit{w/o OPD}.
Although \textit{w/o OPD} reaches 100\% windowed success in one run, it shows greater variability across the two runs and requires more human takeovers.
Specifically, OPD reduces the mean number of human interventions per training run from 52.5 to 20, a reduction of 61.9\%.

We additionally compare \methodname with a conventional pipeline that directly fine-tunes a compact policy using BC on human demonstrations, evaluating both training efficiency and final task success.
For \textit{Pick Fruit}, we collect 120 in-distribution human demonstration trajectories from the target task setting and use them to train the compact policy.
Data collection and training take 94~min in total, and the resulting policy succeeds in 7 of 27 evaluation trials (25.93\%), substantially below the final evaluation success rate of 88.89\% (24/27) achieved by \methodname.
As reported in Table~\ref{tab:success_rate}, \methodname requires 1.003~h (approximately 60.2~min) for post-training, excluding base-teacher fine-tuning, demonstrating higher task success with less adaptation time when a teacher policy is available.

The gains over direct RL and demonstration-only BC stem from complementary effects of OPD and IPR.
OPD provides dense teacher supervision at student-visited states that fixed demonstrations may not cover, reducing compounding action errors without relying on sparse rewards or initially inaccurate critics~\cite{schmitt2018kickstarting}.
Masked-action alignment supplements this supervision by matching the teacher's responses to changes in visual input.
The resulting initialization may improve the quality and coverage of subsequent rollouts, providing more informative data for IQL's value learning and advantage-weighted policy updates~\cite{kostrikov2022iql}.
IPR further uses human corrections and reward feedback to learn recovery behaviors and improve task performance beyond imitation.

\subsubsection{Effect of Mask Alignment on OPD}
\label{sec:masked_supervision}

This ablation tests whether mask alignment improves OPD by providing teacher supervision under masked visual inputs.
We use the \textit{Click Bell} environment and teacher--student models described in Section~\ref{sec:opd_takeover}.
The student policy is distilled from 32 teacher rollouts.
\methodname (\textit{Ours}) uses mask alignment to match the student's and teacher's action predictions under the same image mask (Eq.~\eqref{eq:opd_mask_loss}).
\textit{OPD w/o align} removes this loss, while \textit{Vision Feature Alignment} serves as an alternative alignment baseline.
In \textit{Vision Feature Alignment}, we pool the vision--language model (VLM) features into a 2048-dimensional vector.
A trainable multilayer perceptron (MLP) then projects this vector to the feature dimension of the student's DINOv3 visual encoder.
All three methods are evaluated over three runs of 100 online interactions.
The success rate is computed using a sliding window of 20 interactions.

\begin{figure}[!t]
    \centering
    \includegraphics[width=\columnwidth]{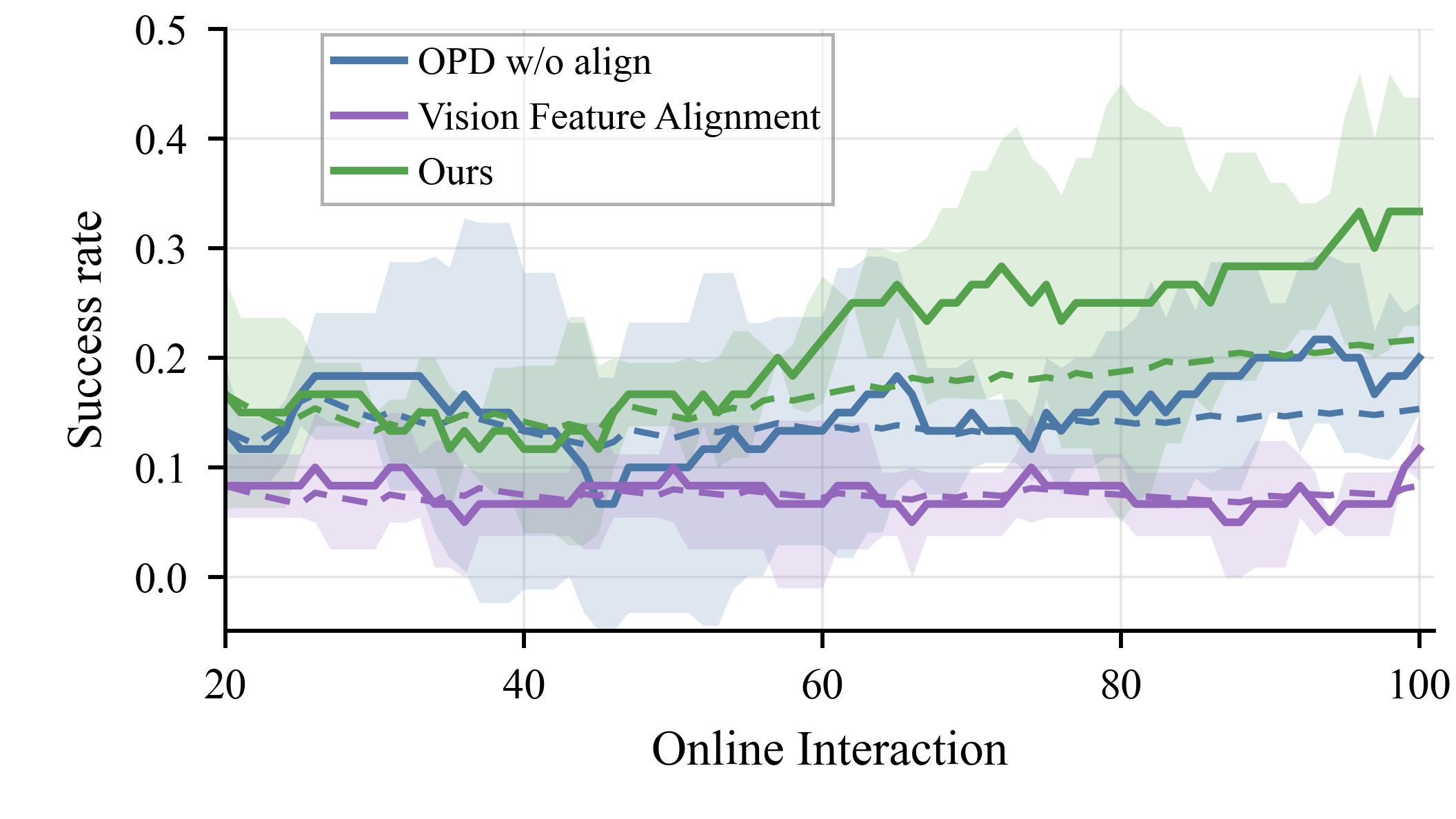}
    \caption{Effect of mask alignment on online OPD.
    Solid curves show the mean trailing-20-interaction success rate across three runs.
    Dashed curves indicate the cumulative success rate.
    Shaded bands indicate the mean $\pm 1$ sample standard deviation across runs.}
    \label{fig:feature_alignment_success}
\end{figure}

Mask alignment achieved the highest mean success rate over the final 20 interactions (Fig.~\ref{fig:feature_alignment_success}).
It reached 33.3\%, compared with 20.0\% for \textit{OPD w/o align} and 11.7\% for \textit{Vision Feature Alignment}.
These results suggest that masked-action alignment helps the student policy focus on image regions that guide the teacher's decisions, thereby improving task success.
\textit{Vision Feature Alignment} can impair OPD performance relative to \textit{OPD w/o align}.
The substantial mismatch between VLM and DINOv3 representations can cause direct alignment to disrupt pretrained visual capabilities.
These findings favor aligning the policies' action responses to image masking over matching their visual features in this setting.

\subsubsection{Effect of Reinforcement Learning in IPR}
\label{sec:ipr_online_bc}

This ablation evaluates the contribution of reinforcement learning within IPR on the real-world \textit{Pick Fruit} and \textit{Hang Cup} tasks.
We construct an Online BC baseline by removing IPR's RL components, including value function learning and advantage weighting.
Online BC learns from the full collected trajectories, including both autonomous policy rollouts and human corrective segments, and directly updates the student using an unweighted BC loss.
Both methods start from the same post-OPD student checkpoint and use the same number of environment interaction episodes for each task.

As shown in Table~\ref{tab:training_progress}, Online BC improves success from 59.26\% after OPD to 66.67\% (18/27) on \textit{Pick Fruit}, and from 25.00\% to 65.00\% (13/20) on \textit{Hang Cup}.
These results show that learning from additional online trajectories and human corrections can improve the student policy.
With RL-based updates, IPR reaches 88.89\% (24/27) and 80.00\% (16/20), exceeding Online BC by 22.22 and 15.00 percentage points, respectively, under the same interaction budget.

These gains are consistent with IPR's use of advantage-weighted regression to reduce the influence of low-quality actions (Eq.~\eqref{eq:ipr_actor_loss}).
Actions with low estimated advantage receive smaller regression weights, reducing their contribution to policy updates.
Online BC instead learns from all recorded actions without weighting them by estimated return, so erroneous autonomous behaviors can be reinforced alongside useful human corrections.

\subsection{Post-training with Agentic and WAM Teachers}
\label{sec:agentic_wam_teachers}
To evaluate \methodname across teacher architectures, we use an agentic teacher and a world action model (WAM) on simulated \textit{Push-T}, which requires pushing a T-shaped object into a target pose.
Following GPT-GEAR~\cite{zhou2026largemultimodalmodelsdirect}, the agentic teacher uses GPT-6.1 Sol with \texttt{low} reasoning effort and selects pushing actions through a tool interface using observations and execution feedback.
The WAM teacher is Fast-WAM~\cite{yuan2026fastwam}, finetuned on 150 task trajectories.
Both teachers remain frozen and supervise their students through action outputs only.

Training and evaluation use the same ten fixed initial configurations.
Each configuration is evaluated four times, yielding 40 episodes per policy, each limited to 400 simulation steps.
We report mean maximum coverage across these episodes.

Each student uses 40 teacher trajectories for offline distillation, followed by 40 OPD and 100 IPR episodes, with an action-chunk length of 16 throughout training and deployment.
Automated takeover is triggered by either stagnation, when neither geometric error nor coverage improves meaningfully over a recent window, or regression, when coverage remains substantially below its recent best value for a sustained period.
During IPR, a planning-based policy supplies corrective actions in place of human interventions.

\begin{table}[t]
    \caption{Mean maximum coverage on Push-T per policy.}
    \label{tab:agentic_pusht}
    \centering
    \footnotesize
    \renewcommand{\arraystretch}{1.2}
    \setlength{\tabcolsep}{2pt}
    \begin{tabular*}{\columnwidth}{@{\extracolsep{\fill}} l c c c c @{}}
        \toprule
        \textbf{Teacher} & \textbf{Base Policy} & \textbf{Offline Dist.} & \textbf{OPD} & \textbf{IPR} \\
        \midrule
        GPT-6.1 Sol & 89.07\% & 49.48\% & 71.29\% & \textbf{95.01\%} \\
        Fast-WAM & 29.24\% & 13.36\% & 18.08\% & \textbf{82.09\%} \\
        \bottomrule
    \end{tabular*}
\end{table}

Table~\ref{tab:agentic_pusht} reports mean maximum coverage after each training stage.
For the GPT teacher, OPD raises student coverage from 49.48\% after offline distillation to 71.29\%, narrowing the gap to the teacher's 89.07\%.
IPR further improves coverage to 95.01\%, exceeding the teacher by 5.94 percentage points.

For Fast-WAM, student coverage increases from 13.36\% after offline distillation to 18.08\% after OPD, remaining below the teacher's 29.24\%.
IPR raises coverage to 82.09\%, showing that \methodname can refine a student initialized from a teacher with limited task performance.

Together with the real-world experiments using GR00T N1.7, these results support the applicability of \methodname across teacher architectures.
Independent student deployment also reduces mean rollout time from 905.8~s for the remotely hosted GPT-6.1 Sol teacher to 1.78~s, an approximately \textbf{\boldmath$509\times$ speedup} that removes teacher-response latency.

\subsection{Effectiveness of RAPID in Various Tasks}

\begin{figure}[t]
    \centering
    \includegraphics[width=\columnwidth]{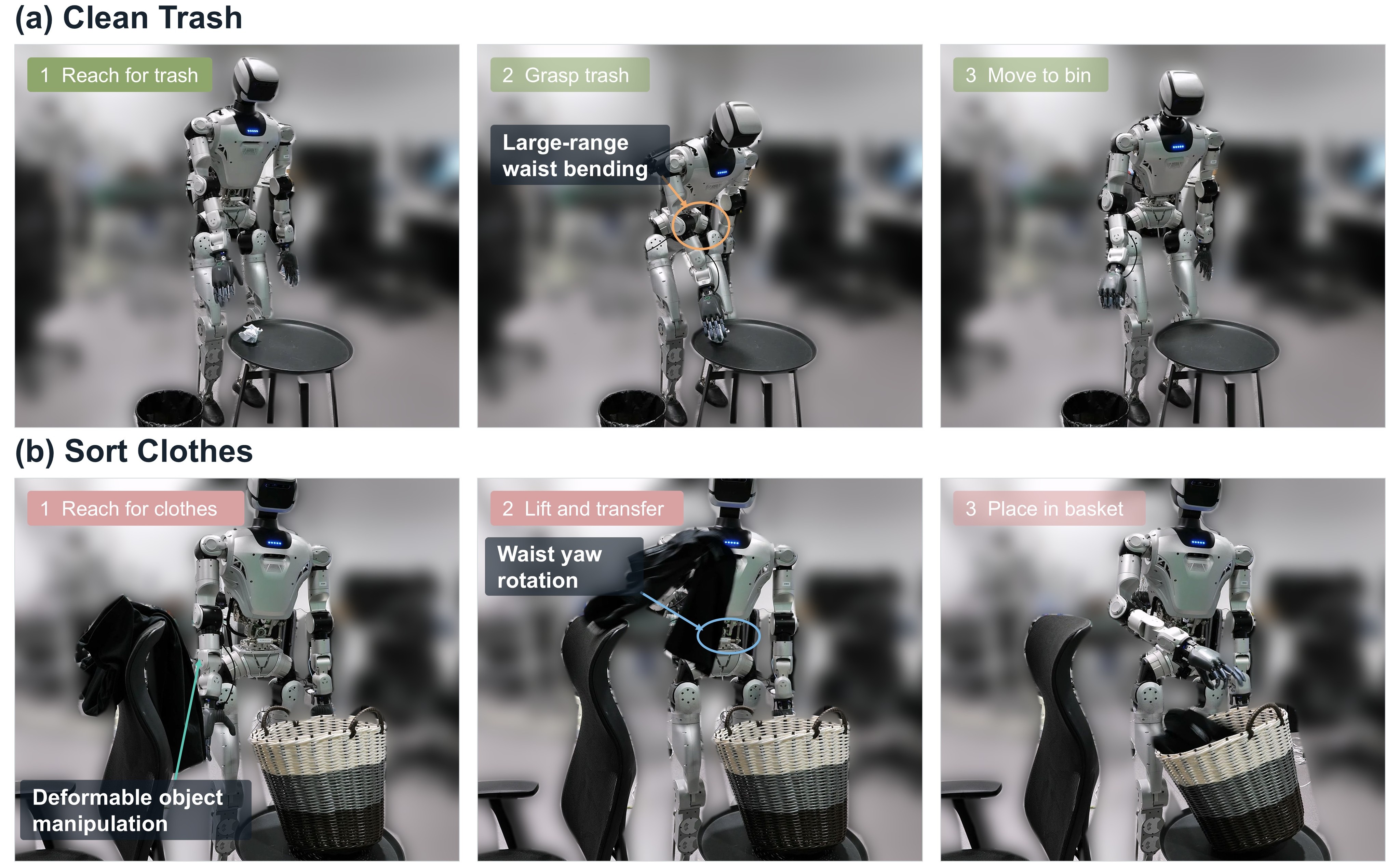}
    \caption{Illustrations of the two additional real-world tasks.
    (a) The robot grasps trash from a table and moves it to a bin, requiring waist bending and squatting.
    (b) The robot transfers clothes from the back of a chair into a basket, involving torso rotation and deformable object manipulation.
    Each row shows successive stages from left to right.}
    \label{fig:additional_tasks}
\end{figure}

The preceding real-world experiments focused primarily on \textit{Pick Fruit} and \textit{Hang Cup}.
To further assess the effectiveness of \methodname across different manipulation tasks, we extend the evaluation to \textit{Clean Trash} and \textit{Sort Clothes} (Fig.~\ref{fig:additional_tasks}).
In \textit{Clean Trash}, the robot must bend at the waist and squat to pick up trash from a tabletop, then place it in a bin.
In \textit{Sort Clothes}, the robot retrieves clothing draped over the back of a chair and places it in a basket.
The training data composition for both tasks is summarized in Table~\ref{tab:training_data}, with the allocation across training stages detailed in the Appendix.

\methodname also improved success rates on the two additional tasks (Table~\ref{tab:additional_tasks}).
Success rates increased from 20\% to 90\% on \textit{Clean Trash} and from 10\% to 100\% on \textit{Sort Clothes}.
The total training times were 0.94 hours for \textit{Clean Trash} and 0.73 hours for \textit{Sort Clothes}.
Across all four real-world tasks, \methodname increases the equally weighted mean success rate from \textbf{20.51\% to 89.72\% (+69.21 percentage points)} within approximately one hour per task on a single consumer GPU.

\begin{table}[t]
    \caption{Initial and final success rates (SR) across all four real-world tasks.}
    \label{tab:additional_tasks}
    \centering
    \footnotesize
    \renewcommand{\arraystretch}{1.2}
    \setlength{\tabcolsep}{2pt}
    \begin{tabular*}{\columnwidth}{@{\extracolsep{\fill}} l c c c c @{}}
        \toprule
        \textbf{Metric} & \textbf{\textit{Pick Fruit}} & \textbf{\textit{Hang Cup}} & \textbf{\textit{Clean Trash}} & \textbf{\textit{Sort Clothes}} \\
        \midrule
        \textbf{Initial SR} & 37.04\% (10/27) & 15.00\% (3/20) & 20\% (4/20) & 10\% (2/20) \\
        \textbf{Final SR} & \textbf{88.89\% (24/27)} & \textbf{80.00\% (16/20)} & \textbf{90\% (18/20)} & \textbf{100\% (20/20)} \\
        \bottomrule
    \end{tabular*}
\end{table}

%% file: sec/conclusion.tex
\section{Conclusion and Future Work}

We presented \methodname, an architecture-agnostic framework for rapid adaptation of pretrained humanoid policies through distillation and online RL.
Offline distillation and OPD with masked-action supervision transfer behaviors from a frozen teacher policy into a compact, task-specific student.
IPR further improves the student through reward feedback and human corrective interventions.
The framework requires zero access to the teacher's parameters or internal features and yields a student policy that runs independently of the teacher.
Across four real-world humanoid loco-manipulation tasks, roughly one hour of post-training per task on a single consumer GPU elevates mean success rates from 20.51\% to 89.72\% (+69.21 percentage points) while cutting required human interventions by 61.9\%.
At runtime, base-free student execution slashes inference compute by 98.68\% relative to generalist VLA backbones and accelerates cloud-hosted agentic policies by $509\times$ while surpassing their task performance.

By decoupling real-time on-robot control from heavy, proprietary foundation architectures, \methodname establishes a scalable bridge between generalist robot pretraining and precision physical deployment.
Building upon this foundation, promising avenues for future research include composing specialized \methodname students into a unified, language-routed whole-body skill library and scaling parallel post-training across humanoid fleets for continual open-world evolution.

%% file: sec/theory_appendix.tex
\subsection{Proof and Interpretation of the Policy Adaptation Bound}
\label{app:unified_adaptation}

\paragraph{Decision process and reference policy}
We use the finite-horizon process and notation from Sec.~\ref{sec:rapid_system}, with $T$ policy decisions and a common initial-state distribution, dynamics, execution rule, and task reward.
For the teacher $\pi_{\mathrm{teacher}}$, define the task-reward value functions
\[
\begin{aligned}
    V_t^{\mathrm{teacher}}(s)
        &=\mathbb{E}_{\mathbf{A}^{T}\mid s}
          Q_t^{\mathrm{teacher}}(s,\mathbf{A}^{T}),\\
    Q_t^{\mathrm{teacher}}(s,\mathbf{A})
        &=\mathbb{E}\bigl[r_t+V_{t+1}^{\mathrm{teacher}}(s_{t+1})\mid s_t=s,\mathbf{A}_t=\mathbf{A}\bigr],
\end{aligned}
\]
with $V_{T+1}^{\mathrm{teacher}}=0$.

\paragraph{Performance difference}
We first express the teacher--student return gap as a sum of one-decision value differences at student-visited states.
The difference $V_t^{\mathrm{teacher}}(s_t)-Q_t^{\mathrm{teacher}}(s_t,\mathbf{A}^{S}_t)$ measures the value lost by choosing the student action for one decision and following the teacher thereafter.
Applying the Bellman equation along a student trajectory, with $\mathbf{A}^{S}_t=\pi_\phi(s_t)$, gives
\begin{equation}
\begin{aligned}
    &\sum_{t=1}^{T}\mathbb{E}_{\pi_\phi}
      \bigl[V_t^{\mathrm{teacher}}(s_t)-Q_t^{\mathrm{teacher}}(s_t,\mathbf{A}^{S}_t)\bigr]\\
    &\quad=\sum_{t=1}^{T}\mathbb{E}_{\pi_\phi}
      \bigl[V_t^{\mathrm{teacher}}(s_t)-r_t-V_{t+1}^{\mathrm{teacher}}(s_{t+1})\bigr]\\
    &\quad=\mathbb{E}_{s_1}[V_1^{\mathrm{teacher}}(s_1)]-J(\pi_\phi)\\
    &\quad=J(\pi_{\mathrm{teacher}})-J(\pi_\phi).
\end{aligned}
\label{eq:app_performance_difference}
\end{equation}
The intermediate value terms cancel across consecutive decisions, leaving only $\mathbb{E}[V_1^{\mathrm{teacher}}(s_1)]$, since $V_{T+1}^{\mathrm{teacher}}=0$.
The common initial-state distribution identifies this remaining term with $J(\pi_{\mathrm{teacher}})$.
This is the finite-horizon performance-difference argument underlying continuous-action imitation analysis~\cite{pan2018agile}.
We next control each value difference by the action-fitting error at the same state.
Using the definition of $V_t^{\mathrm{teacher}}$ and the assumed $L$-Lipschitz continuity of $Q_t^{\mathrm{teacher}}(s,\cdot)$ yields
\[
\begin{aligned}
    &V_t^{\mathrm{teacher}}(s)-Q_t^{\mathrm{teacher}}(s,\mathbf{A}^{S})\\
    &=\mathbb{E}_{\mathbf{A}^{T}\mid s}
      \bigl[Q_t^{\mathrm{teacher}}(s,\mathbf{A}^{T})-Q_t^{\mathrm{teacher}}(s,\mathbf{A}^{S})\bigr]\\
    &\le L\,\mathbb{E}_{\mathbf{A}^{T}\mid s}\|\mathbf{A}^{S}-\mathbf{A}^{T}\|_2\\
    &=L\,\ell_{\mathrm{BC}}(s,\phi).
\end{aligned}
\]
Substitution into Eq.~\eqref{eq:app_performance_difference} gives
\begin{equation}
\begin{aligned}
    &J(\pi_{\mathrm{teacher}})-J(\pi_\phi)\\
    &\quad\le L\sum_{t=1}^T\mathbb{E}_{s\sim d_t^{\pi_\phi}}\ell_{\mathrm{BC}}(s,\phi)\\
    &\quad=LT\,\mathbb{E}_{s\sim d^{\pi_\phi}}\ell_{\mathrm{BC}}(s,\phi).
\end{aligned}
\label{eq:app_on_policy_action_bound}
\end{equation}
The explicit factor $T$ follows from using cumulative reward together with a normalized, time-averaged state distribution.

\paragraph{Transferring between state distributions}
The preceding bound involves the deployment distribution $d^{\pi_\phi}$, whereas $\epsilon_{\mathrm{fit}}$ is defined under the supervision distribution $q$.
We therefore bound the change in expected action error between these distributions.
Set $f(s)=\ell_{\mathrm{BC}}(s,\phi)\in[0,B]$.
For $B>0$, write $f(s)=\int_0^B\mathbf{1}\{f(s)>z\}\,\mathrm{d}z$.
Taking expectations under the two distributions and subtracting yields
\[
\begin{aligned}
    \mathbb{E}_{d^{\pi_\phi}}f-\mathbb{E}_{q}f
    &=\int_0^B\!\left[
       \Pr_{d^{\pi_\phi}}(f>z)-\Pr_q(f>z)\right]\,\mathrm{d}z\\
    &\le B\,\mathrm{TV}(d^{\pi_\phi},q).
\end{aligned}
\]
For each $z$, the set $\{s:f(s)>z\}$ is an event, so its probability difference is at most $\mathrm{TV}(d^{\pi_\phi},q)$; integration over an interval of length $B$ gives the inequality.
The case $B=0$ is immediate.
Under our convention, $\mathrm{TV}(p,q)=\tfrac12\|p-q\|_1$ when densities exist, so the coefficient is $B$, rather than $2B$.
Recalling $\epsilon_{\mathrm{fit}}=\mathbb{E}_{s\sim q}\ell_{\mathrm{BC}}(s,\phi)$, we obtain
\[
    \mathbb{E}_{s\sim d^{\pi_\phi}}\ell_{\mathrm{BC}}(s,\phi)
    \le \epsilon_{\mathrm{fit}}+B\,\mathrm{TV}(d^{\pi_\phi},q).
\]
Substituting this inequality into Eq.~\eqref{eq:app_on_policy_action_bound} controls the student's return shortfall relative to the teacher:
\[
\begin{aligned}
    &J(\pi_{\mathrm{teacher}})-J(\pi_\phi)\\
    &\quad\le LT\bigl[\epsilon_{\mathrm{fit}}+B\,\mathrm{TV}(d^{\pi_\phi},q)\bigr].
\end{aligned}
\]
Finally, to compare the student with the optimal policy, add the teacher's own shortfall $\Delta_{\mathrm{teacher}}=J(\pi^*)-J(\pi_{\mathrm{teacher}})$:
\[
\begin{aligned}
    &J(\pi^*)-J(\pi_\phi)\\
    &\quad=\bigl[J(\pi^*)-J(\pi_{\mathrm{teacher}})\bigr]
      +\bigl[J(\pi_{\mathrm{teacher}})-J(\pi_\phi)\bigr]\\
    &\quad\le LT\Bigl[
      \epsilon_{\mathrm{fit}}
      +B\,\mathrm{TV}(d^{\pi_\phi},q)
      \Bigr]
      +\Delta_{\mathrm{teacher}}.
\end{aligned}
\]
This is precisely Eq.~\eqref{eq:unified_adaptation_bound}, completing the proof of Theorem~\ref{thm:unified_adaptation}.
\hfill$\square$